\documentclass[10pt,journal,compsoc]{IEEEtran}
\usepackage{url}
\usepackage{graphicx}
\usepackage{multirow}
\usepackage{multicol}
\usepackage{amsmath,amssymb}
\usepackage{subfigure}
\usepackage{color}
\usepackage{ulem}
\usepackage{xcolor}
\usepackage{color}
\usepackage{hyperref} 
\usepackage{epsfig}
\usepackage{float} 
\newenvironment{myitemize}[1][]{
	\begin{list}{$\bullet$}
		{
			\setlength{\leftmargin}{5mm}
			\setlength{\parsep}{1mm}
			\setlength{\topsep}{0mm}
			\setlength{\itemsep}{0mm}
			\setlength{\labelsep}{1.5mm}
			\setlength{\itemindent}{0mm}
			\setlength{\listparindent}{5mm}
	}}
	{\end{list}}
\usepackage[justification=centering]{caption}
\ifCLASSOPTIONcompsoc
  \usepackage[nocompress]{cite}
\else
  \usepackage{cite}
\fi
\ifCLASSINFOpdf
\else
\fi
\begin{document}
%
\title{Large Scale Visual Food Recognition}
\author{Weiqing Min,~\IEEEmembership{Member,~IEEE,}
         Zhiling Wang,  Yuxin Liu, Mengjiang Luo, Liping Kang, Xiaoming Wei, Xiaolin Wei,
         Shuqiang Jiang,~\IEEEmembership{Senior Member,~IEEE.}
\IEEEcompsocitemizethanks{
\IEEEcompsocthanksitem Weiqing Min, Zhiling Wang, Yuxin Liu, Mengjiang Luo, Shuqiang Jiang are with the Key Laboratory of Intelligent Information
Processing, Institute of Computing Technology, Chinese Academy of Sciences, Beijing 100190, China, and also with University of Chinese Academy of Sciences, Beijing, 100049, China.\protect\\
E-mail: minweiqing@ict.ac.cn, \{zhiling.wang, yuxin.liu,  mengjiang.luo\}@vipl.ict.ac.cn, sqjiang@ict.ac.cn.
\IEEEcompsocthanksitem Liping Kang, Xiaoming Wei, Xiaolin Wei are from Meituan Group\protect\\
E-mail: \{kangliping, weixiaoming, weixiaolin02\}@meituan.com.
}
}

\IEEEtitleabstractindextext{%
\begin{abstract}
Food recognition plays an important role in food choice and intake, which is  essential to the health and well‐being of humans.  It is thus of  importance  to the computer vision community, and can further support many food-oriented vision and multimodal tasks, e.g., food detection and segmentation, cross-modal recipe retrieval and generation. Unfortunately, we have witnessed remarkable advancements in generic visual recognition for  released large-scale datasets, yet largely lags in the food domain. In this paper, we introduce Food2K, which is the largest food recognition dataset with 2,000 categories and over 1 million images. Compared with existing food recognition datasets, Food2K bypasses them in both categories and images by one order of magnitude, and thus establishes a new challenging benchmark to develop advanced models for food
visual representation learning. Furthermore, we propose a deep progressive region enhancement network for food recognition, which mainly consists of two components, namely progressive local feature learning  and region feature enhancement. The former adopts improved progressive training to learn diverse and complementary local features, while the latter utilizes self-attention to incorporate richer context with multiple scales into local features
for further local feature enhancement. Extensive experiments on Food2K demonstrate the effectiveness of our proposed method.  More importantly, we have verified better generalization ability of Food2K in various tasks, including food image recognition, food image retrieval, cross-modal recipe retrieval, food detection and segmentation. Food2K  can be further explored to benefit  more food-relevant  tasks including emerging and more complex ones (e.g., nutritional understanding of food), and  the trained models on Food2K can  be expected as  backbones to improve the performance of more food-relevant tasks. We also hope Food2K can serve as a large scale fine-grained visual recognition benchmark, and contributes to the development of large scale fine-grained visual analysis. The dataset, code  and models
are publicly available at \url{http://123.57.42.89/FoodProject.html}.
\end{abstract}

\begin{IEEEkeywords}
Food dataset, food recognition, large-scale datasets, fine-grained recognition.
\end{IEEEkeywords}}

\maketitle
\IEEEdisplaynontitleabstractindextext
\IEEEpeerreviewmaketitle

\IEEEraisesectionheading{\section{Introduction}\label{sec:introduction}}

Food computing~\cite{Min2019A} recently has  come into the focus of public attention as one new emerging area for  supporting  many food-related issues, e.g., food choice~\cite{Boswell-FC-PNAS2018} and healthy diets~\cite{David-DH-Nature2014}. As one basic task in food computing, food recognition  plays an important role for humans in identifying food for further food gathering to meet their survival needs~\cite{Rozin-SFRHO-SB1976}. It is also an essential step in many health applications, such as nutritional understanding of food~\cite{Meyers-Im2Calories-ICCV2015,Quin-Nutrition5k-CVPR2021} and dietary management~\cite{Lu-AIS-TMM2020}.  In addition, food image recognition is an important branch of fine-grained visual classification, and thus has important theoretical research significance. For these reasons, food recognition  has been drawing more  attention in  computer vision and beyond \cite{Bossard-Food101-ECCV2014,Qiu-MDFR-BMVC2019,Martinel-WSR-WACV2018,Parneet-FoodX251-CVPRW2019,Min-MSMVFA-TIP2019,Lixi2019Mixed,Zhao-FL-WACV2021}.

Existing works mainly focus on utilizing medium-scale or small-scale image datasets for food recognition, such as ETH Food-101~\cite{Bossard-Food101-ECCV2014}, Vireo Food-172~\cite{Chen-DIRCRR-MM2016} and ISIA Food-500~\cite{Min-ISIA-500-MM2020}.  They are probably insufficient to build more complicated and advanced statistical models for food computing due to their inadequate  food categories and images. Considering that  large-scale datasets have  been key enablers of  progress in  general image classification and understanding, a large-scale food image dataset is urgently needed  for developing advanced  food visual  representation learning algorithms, and can further  support various food-relevant tasks, such as cross-modal recipe retrieval and generation~\cite{Salvador-LCME-arXiv2018,WangLHM20-SAGN-ECCV2020}.

To this end, we introduce a new large-scale  benchmark dataset Food2K for food recognition as the main contribution. Food2K contains 1,036,564 images with 2,000 categories, belonging to different super-classes, such as vegetables, meat, barbecue and fried food. In contrast to  existing  datasets, such as ETH Food-101, the size of Food2K in both categories and images bypasses their size by one order of magnitude. In addition to the size, we have carried out rigorous data cleaning, iterative annotation and multiple professional inspection to guarantee its high quality. We hope that this large scale, high-quality dataset will be a useful resource for developing advanced food image representation learning and understanding methods to support both popular and emerging food-relevant vision tasks. In addition, Food2K can be expected to be one large-scale fine-grained visual recognition benchmark to enable the development of fine-grained visual recognition.

Based on this dataset, we propose a deep progressive region enhancement network for food recognition. It can jointly learn diverse and complementary local and global features for food recognition. We adopt the  progressive training strategy to obtain abundant and various food features from food images, like different ingredient information. This strategy can help  to learn comprehensive and multiple fine-grained information as training progresses. In addition, our model  incorporates richer context with multiple scales into local features via self-attention to enhance local feature representation. 

Extensive experiments on Food2K demonstrate the effectiveness of our proposed method. In addition, we  provide extensive experiments comparing various state-of-the-art methods for image representation learning, including popular deep  networks, fine-grained methods and existing food recognition methods. Furthermore, we also show that  the networks  learned on Food2K can benefit various food-relevant vision tasks, i.e.,  food recognition, food image retrieval, cross-modal recipe retrieval, food detection and segmentation, indicating better generalization ability of Food2K. The developed networks on Food2K can be expected as the backbone to support more food-relevant vision tasks, especially emerging and more complex ones.
 
The contributions of our paper can be summarized as follows:
\begin{myitemize}
 \item We introduce a new large-scale high-quality food  recognition benchmark  Food2K, which  is the largest food image dataset  with 2,000 categories and 1,036,564 images. 
 \item We propose a deep progressive region enhancement network to learn food-oriented  local  features by  progressive training, and  further utilize self-attention to enhance local features for food recognition. 
 \item We conduct extensive evaluation on Food2K to verify the effectiveness of our approach, where  extensive baselines on this benchmark are provided, including popular deep networks, fine-grained recognition methods and existing food recognition ones. 
 \item We explore the ability of models trained on Food2K to transfer to various food-relevant tasks including  visual food recognition, retrieval, detection, segmentation and cross-modal recipe retrieval, and demonstrate its better generality of Food2K on these tasks.  
\end{myitemize}

\section{Related Work}
\begin{table}[!t]
	\caption{Comparison of current food recognition datasets.}
	\begin{center}
		\begin{tabular}{ccccc}
			\hline
			Dataset& Year&Classes/Images&Type&Public\\
			\hline
			PFID~\cite{Chen2009PFID} &2009 &101/4,545&Western&$\times$\\
			Food50~\cite{Joutou2010A}&2010&50/5,000&Misc.&$\times$\\
			Food85~\cite{Hoashi2010Image}&2010&85/8,500&Misc.&$\times$\\
			UEC Food100~\cite{Matsuda2012Multiple} &2012& 100/14,361&Japanese&$\surd$\\
			UEC Food256~\cite{kawano2014automatic} &2014& 256/25,088&Japanese&$\surd$\\
			ETH Food-101~\cite{Bossard-Food101-ECCV2014} &2014& 101/101,000&Western&$\surd$\\
			Diabetes ~\cite{Anthimopoulos2014A}&2014&11/4,868&Misc.&$\times$\\
			UPMC Food-101~\cite{Wang-RRLMFD-ICME2015}&2015&101/ 90,840&Western&$\surd$\\
			Geo-Dish~\cite{XuRuihan-GMDR-TMM2015}&2015&701/117,504&Misc.&$\surd$\\
			UNICT-FD889~\cite{Farinella2015A}&2015&889/3,583&Misc.&$\surd$\\
			Vireo Food-172~\cite{Chen-DIRCRR-MM2016}&2016&172/110,241&Chinese&$\surd$\\
			Food-975~\cite{Zhou_FGIC_CVPR2016}&2016 & 975/37,785&Misc.&$\times$\\
			Food500~\cite{Merler2016} &2016& 508/148,408&Misc.&$\times$\\
			Food11~\cite{Singla-FnFC-MM2016}  &2016&11/16,643&Misc.&$\times$\\
			UNICT-FD1200~\cite{Farinella2016Retrieval} &2016& 1,200/4,754&Misc.&$\surd$\\
			Food524DB~\cite{ciocca2017learning}&2017&524/247,636&Misc.&$\surd$ \\
			ChineseFoodNet~\cite{Chen2017ChineseFoodNet}&2017&208/192,000&Chinese&$\surd$ \\
			Vegfru~\cite{Hou-VegFru-ICCV2017}&  2017&292/160,000&Misc.&$\surd$\\
			Sushi-50~\cite{Qiu-MDFR-BMVC2019}&  2019&50/3,963&Japanese&$\surd$\\
			FoodX-251~\cite{Parneet-FoodX251-CVPRW2019}&  2019&251/158,846&Misc.&$\surd$\\
			ISIA Food-200~\cite{Min-IGCMAN-ACMMM2019}&  2019&200/197,323&Misc.&$\surd$\\
			FoodAI-756~\cite{Doyen-FoodAI-KDD2019}&  2019&756/400,000&Misc.&$\times$\\
			ISIA Food-500~\cite{Min-ISIA-500-MM2020}&  2020&500/399,726&Misc.&$\surd$\\						 
			\hline
			\textbf{Food2K (Ours)}&\textbf{2021}&\textbf{2,000/1,036,564}&\textbf{Misc.}&$\surd$\\
			\hline
		\end{tabular}
	\end{center}
	\label{datasets_summary}
\end{table}
\textbf{Food-Centric Datasets.} Over the years, the size of food-centric datasets has grown steadily. For example, Bossard \textit{et al.}~\cite{Bossard-Food101-ECCV2014} constructs one western food dataset ETH Food-101 with 101,000 images from 101 food categories. VIREO Food-172~\cite{Chen-DIRCRR-MM2016} consists of 110,241 images from 172 Chinese food categories.  Compared with these two datasets, FoodX-251~\cite{Parneet-FoodX251-CVPRW2019} is released with 158,846 images and 251 categories in the Fine-Grained Visual Categorization Challenge held in conjunction with CVPR2019. However, these datasets fail short in terms of both more comprehensive coverage of food categories and large scale of food images, like ImageNet~\cite{Deng-ImageNet-CVPR2009}. Although the full set of ImageNet contains about 1000 food-related categories~\cite{Yanai-FIR-ICMEW2015}, different from existing food datasets and Food2K, which  contains food classes for direct eating, many food-relevant categories from ImageNet belong to nutrient composition (e.g., choline, vitamin), cooking methods (e.g., split, mix), kitchen ware (e.g., mixer, kibble), etc. The reason is that the aim of ImageNet is for generic object recognition, not particularly for food recognition. Considering  the important role of large-scale datasets in the continuous improvement of visual recognition algorithms, especially for deep learning based methods, we build  a large-scale food recognition dataset  Food2K  with more comprehensive  coverage of categories and larger quantity of images. In Table~\ref{datasets_summary}, we give statistics of  existing food recognition benchmarks together with  Food2K. The size of  Food2K in both categories and images bypasses the size of  existing  datasets by at least one order of magnitude. Although there are some datasets, such as UNICT-FD1200 with larger categories, the quantity of images for each category is very limited.

In addition, there are other food-relevant  recipe datasets, such as Yummly66K~\cite{Min-YAWYE-TMM2018} and Recipe1M~\cite{Salvador-LCME-arXiv2018}.  The most known dataset is Recipe1M. Food2K and Recipe1M belong to large-scale food-related datasets, but with two important differences: (1) Recipe1M is used for cross-modal embedding and retrieval between recipes and images while the released Food2K aims at advancing scalable food visual feature learning. (2) Recipe1M mainly contains over 1 million structured cooking recipes, where each recipe is associated with some food images while Food2K contains over 1 million  images, belonging to 2,000 food categories. We believe  Food2K and  Recipe1M are very  complementary and jointly promote the development of visual  analysis and understanding of food.

\textbf{Food Image Analysis.} The availability of more food datasets has further enabled progress in food recognition. More importantly, recognizing food directly from images is  highly desirable for  various food-related applications, such as nutrient assessment~\cite{Meyers-Im2Calories-ICCV2015,Lu-AIS-TMM2020}, food logging~\cite{Doyen-FoodAI-KDD2019} and self-service settlement~\cite{Aguilar2018Grab}. For these reasons, we have seen an explosion in food recognition algorithms in recent years.  

Although food recognition belongs to fine-grained analysis, it has its unique characteristics\cite{Min2019A}. First, food images don't have distinctive spatial layout. A large number of dishes have deformable food appearance and thus lack the rigid structures. Food consists of ingredients. Ingredients from various types of food images are distributed randomly on a plate. There exists the overlap among different ingredients in the same food image. Even the same ingredient may appear distinctly in different food images. Such complex ingredient distributions in the food images make the task different from other ones like scene recognition with distinctive features such as buildings and trees. Second, food image recognition belongs to fine-grained classification, and thus  has the same problem as fine-grained classification, such as subtle differences among different categories. Existing fine-grained classification methods generally focus on discovering the fixed semantic parts as one important part of its  representation. However, the common semantic parts do not exist in many food categories. Therefore, we should re-design the fine-grained categorization method for food recognition.

In the earlier years,  various hand-crafted features, such as color, texture and SIFT are utilized  for food recognition \cite{Yang-FR-CVPR2010,Bossard-Food101-ECCV2014,Martinel2016A}. In the deep learning era,  because of its powerful  capacity of feature representation, more and more works  resort to different deep networks for food recognition \cite{Martinel-WSR-WACV2018,Qiu-MDFR-BMVC2019,Doyen-FoodAI-KDD2019,Parneet-FoodX251-CVPRW2019}. For example, Qiu  \textit{et al.}~\cite{Qiu-MDFR-BMVC2019}  propose a PAR-Net  to mine discriminative food regions to  improve the performance of classification. There are also some recent works on few-shot food recognition~\cite{Zhao-FL-WACV2021,Shuqiang-FSFR-TOMM2020}. For example, Zhao \textit{et al.}~\cite{Zhao-FL-WACV2021} propose a fusion learning framework, which  utilizes a graph convolutional network  to capture inter-class relations between  image representations and semantic embeddings of different categories for both few-shot and many-shot food recognition.  In addition, there are many works~\cite{Beijbom-MeMa-WACV2015,Zhou_FGIC_CVPR2016,Min-IGCMAN-ACMMM2019,XuRuihan-GMDR-TMM2015,Horiguchi-PCFIR-TMM2018}, which introduce additional context information, e.g., GPS and ingredient information to  improve the recognition performance. For example, Zhou \textit{et al.} \cite{Zhou_FGIC_CVPR2016} mine  rich relationships among ingredients  and restaurant information through the bi-partite graph for food recognition. Min \textit{et al.}~\cite{Min-IGCMAN-ACMMM2019} utilize  ingredients as additional supervised signal to localize multiple informative regions and fused these regional features into the final representation for recognition. Different from these works, considering the characteristics of food images, we adopt one  progressive training strategy to  learn comprehensive  and multiple local features, and further utilize self-attention to incorporate richer contexts with multiple scales to enhance local features. Like general object image analysis including recognition, detection and segmentation, in addition to food recognition, there are also some works on food image detection and segmentation towards more accurate nutritional information extraction\cite{Aguilar2018Grab,ege2019new}. However, they are still under slow progress for the lack of food-relevant datasets and complex ingredient distributions in food images.

These above-mentioned food visual analysis methods generally use RGB food images, and thus probably do not achieve satisfactory performance for further  nutritional content prediction since the volume information is not obtained. For that, RGB-D food analysis benchmarks for nutritional evaluation are built\cite{Quin-Nutrition5k-CVPR2021}, where additional depth images can be used to estimate the food volume for the improved prediction of calorie and macronutrients.

\textbf{Multimodal Food Learning.} Multimodal food image-recipe text joint learning is another food related topic in this community. There are mainly two lines for this study. The first line is cross-modal food image-recipe text retrieval \cite{salvador2017learning,wang2019learning,papadopoulos2022learning,fu2020mcen,Salvador-LCME-arXiv2018}, where the main idea is to learn a joint embedding of food images and recipes to support the retrieval between food images and recipes. Due to the complex visual appearance of food images, effective visual feature learning is still one  key. In addition, various types of text annotations should be further considered. The recent work\cite{Salvador-HT-CVPR2021} proposed  a hierarchical transformer to achieve better cross-modal retrieval performance. The other one is cross-modal food image or cooking recipe generation\cite{Papadopoulos-MakePizza-CVPR2019,Fangda-CookGAN-WACV2020,Zhu-CookGAN-CVPR2020,salvador2019inverse,WangLHM20-SAGN-ECCV2020}. For example, Wang \textit{et al.}\cite{WangLHM20-SAGN-ECCV2020} proposed one structure-aware generation network to generate cooking instructions based on only food images and ingredients. All of these works also involve visual feature learning from food images. Constructing large annotated food image datasets can help the development of food visual recognition models, and also supports the multimodal food learning task. 

\textbf{Food Computing.} Food computing~\cite{Min2019A} has raised great interest recently for its various applications in health, culture, etc. It contains different tasks, such as food recognition~\cite{Zhao-FL-WACV2021,Qiu-MDFR-BMVC2019,Parneet-FoodX251-CVPRW2019}, detection~\cite{Aguilar2018Grab}, segmentation~\cite{ege2019new}, retrieval\cite{salvador2017learning,wang2019learning,fu2020mcen} and generation \cite{Salvador-IC-CVPR2019,WangLHM20-SAGN-ECCV2020}. Among these tasks, food recognition  is an important and basic task for further  supporting more complex food-relevant vision and multimodal tasks. Therefore, constructing  large scale food recognition datasets and designing advanced food recognition algorithms on these large-scale datasets are very vital for the development of food computing.
\begin{figure*}[htbp]
	\scriptsize
	\centering
	\subfigure[]{
		\begin{minipage}[t]{0.4\linewidth}
			\centering
			\includegraphics[width=2.8in]{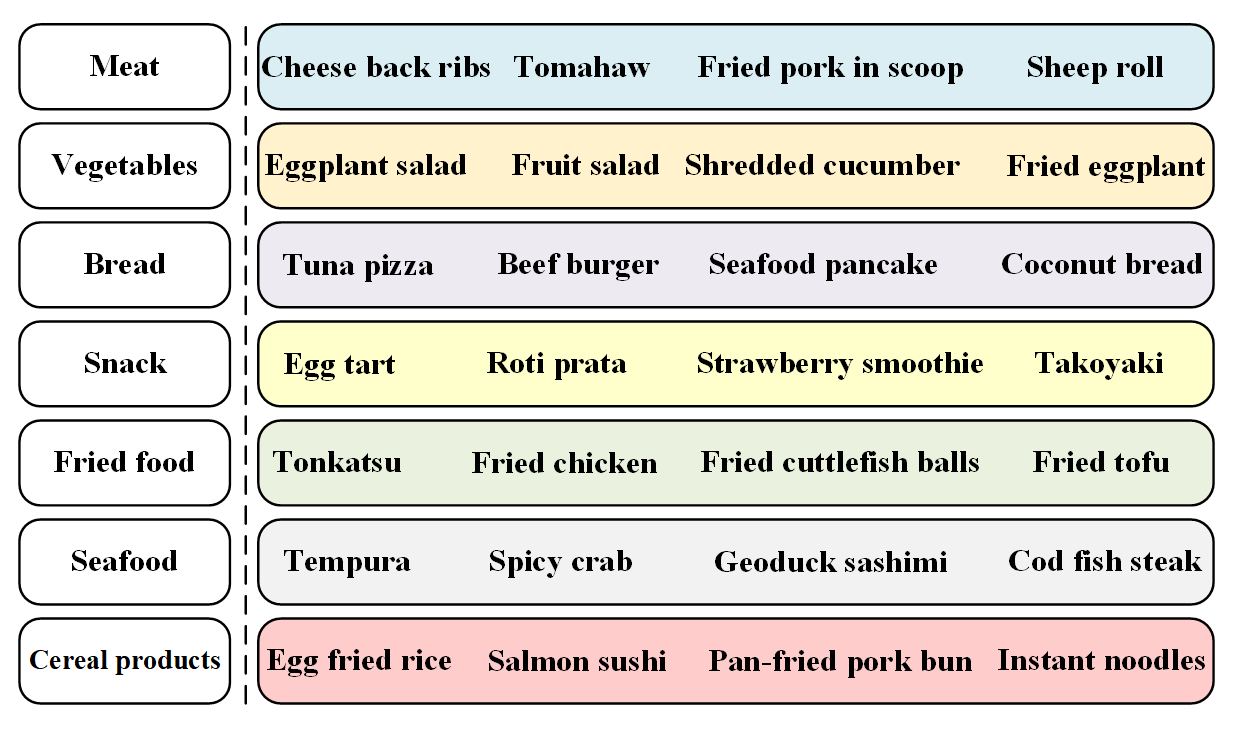}
		\end{minipage}%
	}%
	\subfigure[]{
		\begin{minipage}[t]{0.5\linewidth}
			\centering
			\includegraphics[width=3.9in]{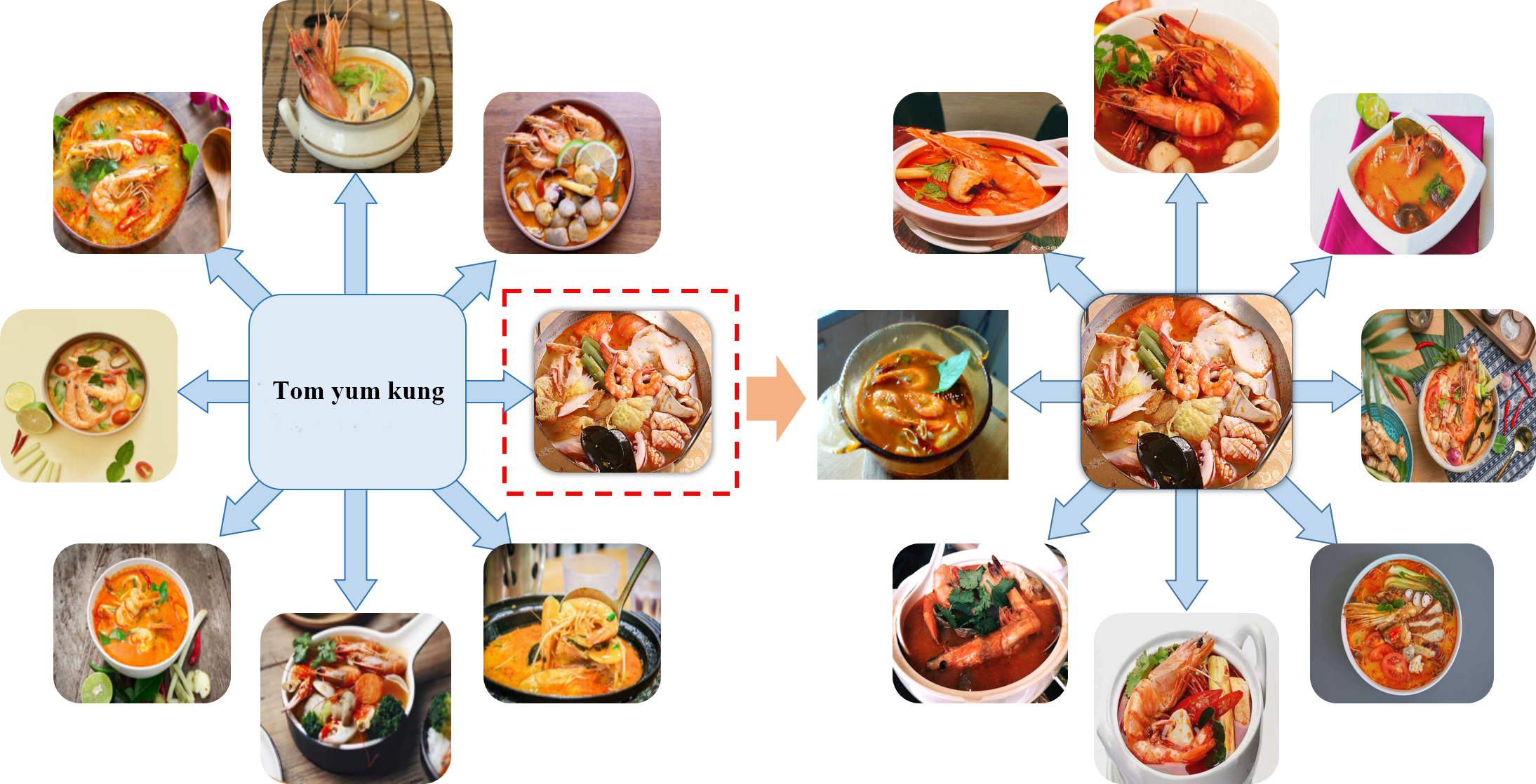}
		\end{minipage}%
	}%
	\centering
	\caption{(a) some categories from Food2K and (b) an example for collecting images via  labels and image based retrieval.}
	\label{Data_Collection_example}
\end{figure*} 

\section{Food-2K Dataset}
\label{ISIA Food-2000_Dataset}
\subsection{Dataset Construction}

\begin{figure}
	\centering
	\includegraphics[width=0.5\textwidth]{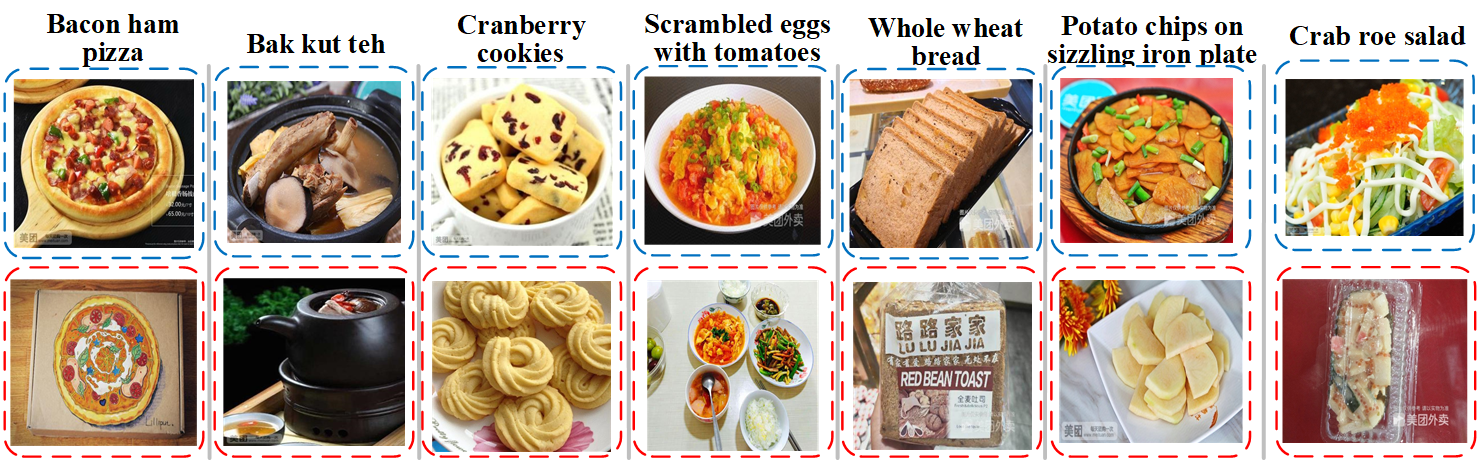}
	\caption{Example annotations for Food2K. For each column, the blue dashed box denotes  the qualified image for this  category, while the  red  one is unqualified. Different types of unqualified ones are showed, e.g., the painting food, the occlusion of the main part, missed important ingredients, and more categories in one image.}
	\label{different_forms}
\end{figure}
\begin{figure}
	\centering
	\includegraphics[width=0.48\textwidth]{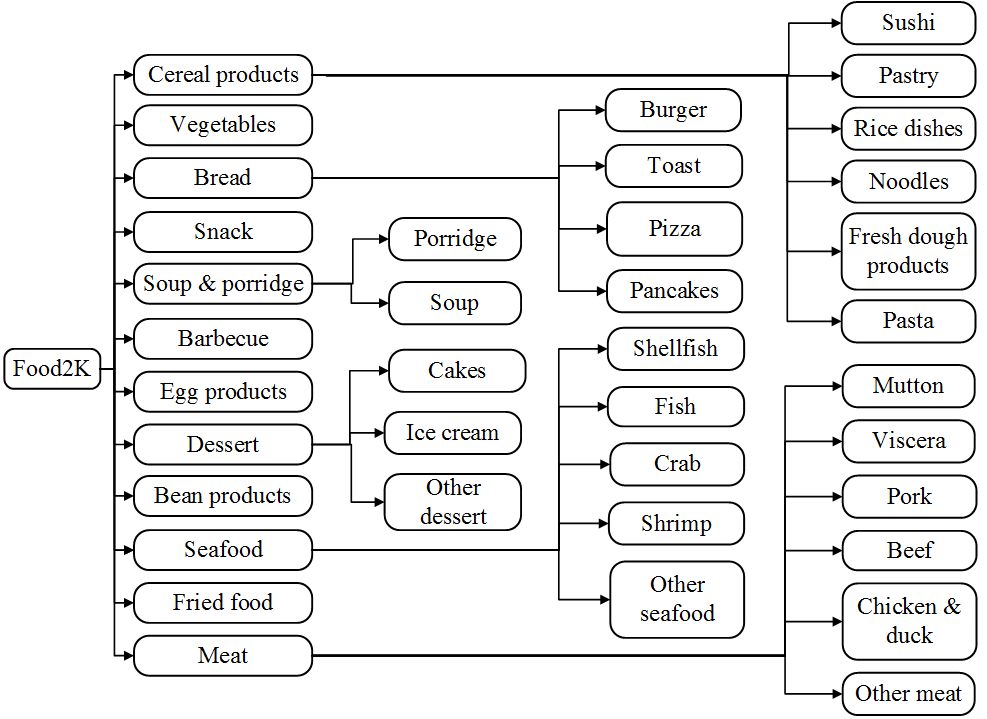}
	\caption{The ontology of Food2K.}
	\label{Hierarchy-tree_ontology}
\end{figure}
\begin{figure}
	\centering
	\includegraphics[width=0.5\textwidth]{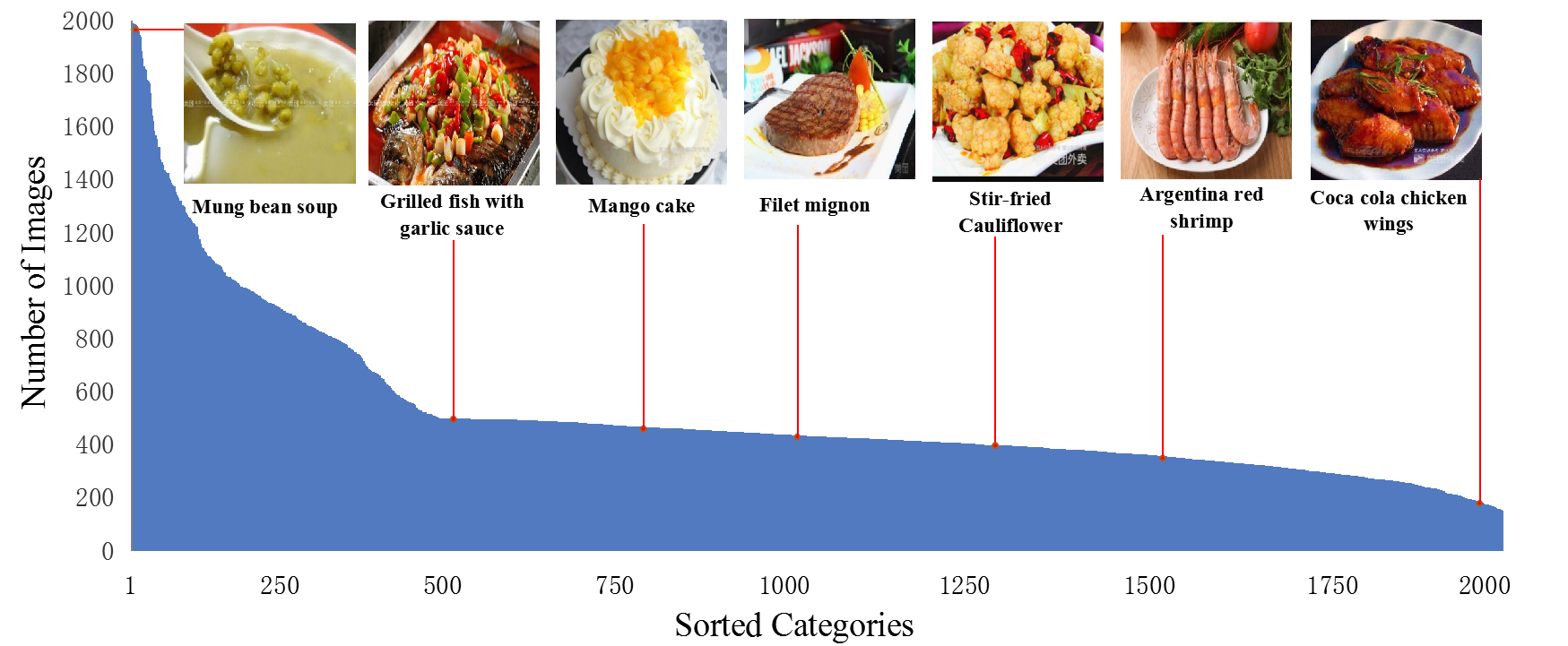}
	\caption{The distributions over each category in the Food2K.}
	\label{distribution2}
\end{figure}
\begin{figure*}[htbp]
	\scriptsize
	\centering
	\subfigure[]{
		\begin{minipage}[t]{0.5\linewidth}
			\centering
			\includegraphics[width=3.5in]{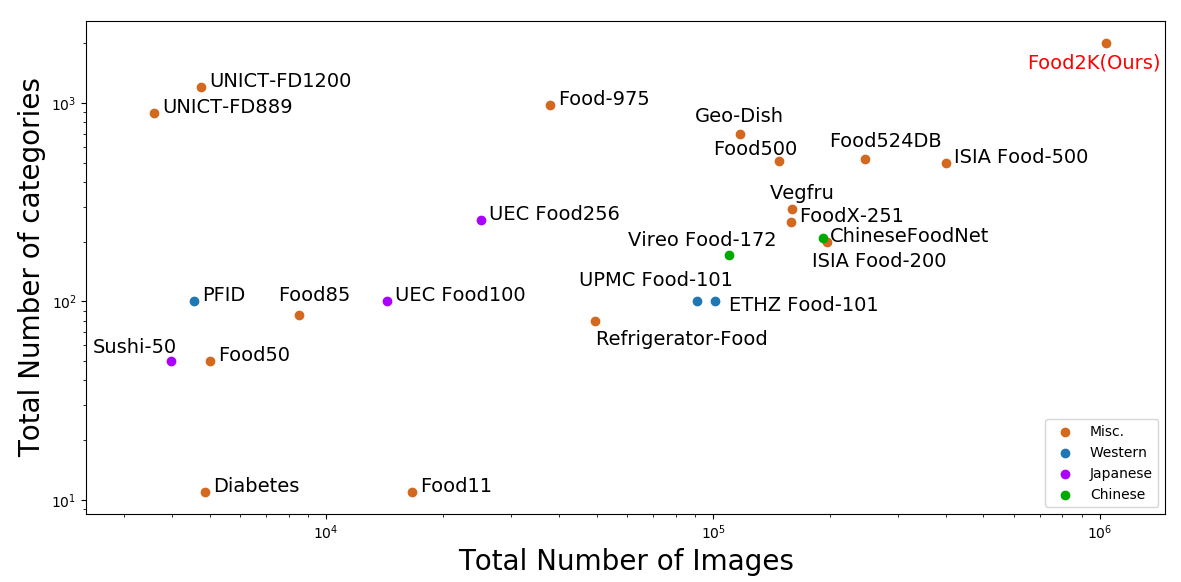} 
		\end{minipage}%
	}%
	\subfigure[]{
		\begin{minipage}[t]{0.5\linewidth}
			\centering
			\includegraphics[width=3.3in]{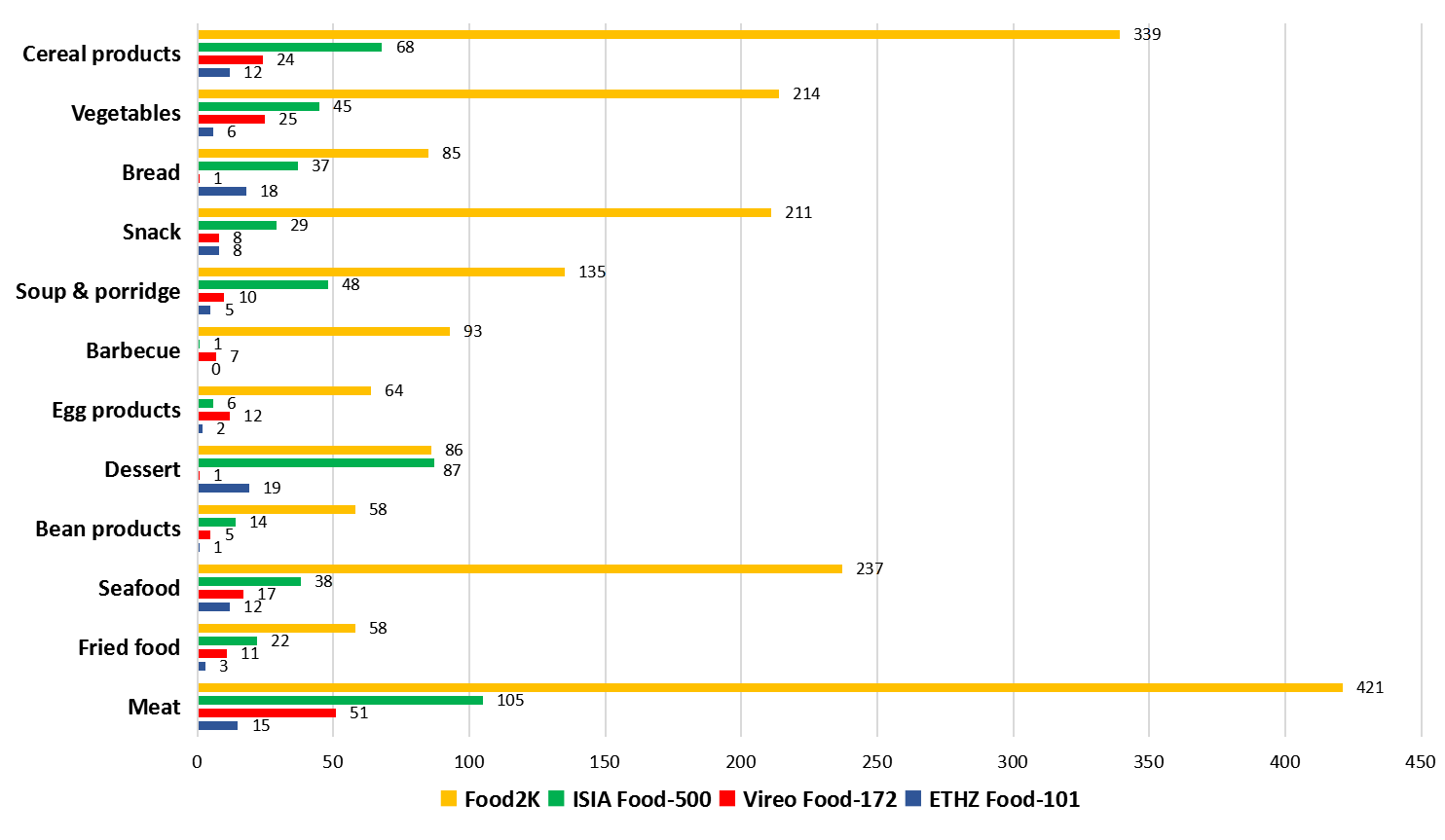}
		\end{minipage}%
	}%
	\centering
	\caption{(a) Datasets  distributed across number of images and categories and (b) distributions of categories on Food2K and typical datasets.~\\}
	\label{Data_Collection_ComP}
\end{figure*}

\begin{figure*}[htbp]
	\scriptsize
	\centering
	\subfigure[]{
		\begin{minipage}[t]{0.5\linewidth}
			\centering
			\includegraphics[width=3.5in]{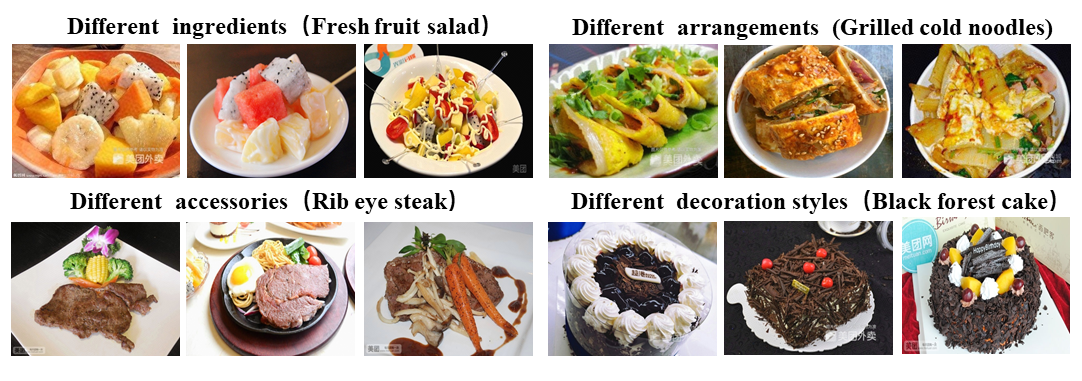}
		\end{minipage}%
	}%
	\subfigure[]{
		\begin{minipage}[t]{0.5\linewidth}
			\centering
			\includegraphics[width=3.5in]{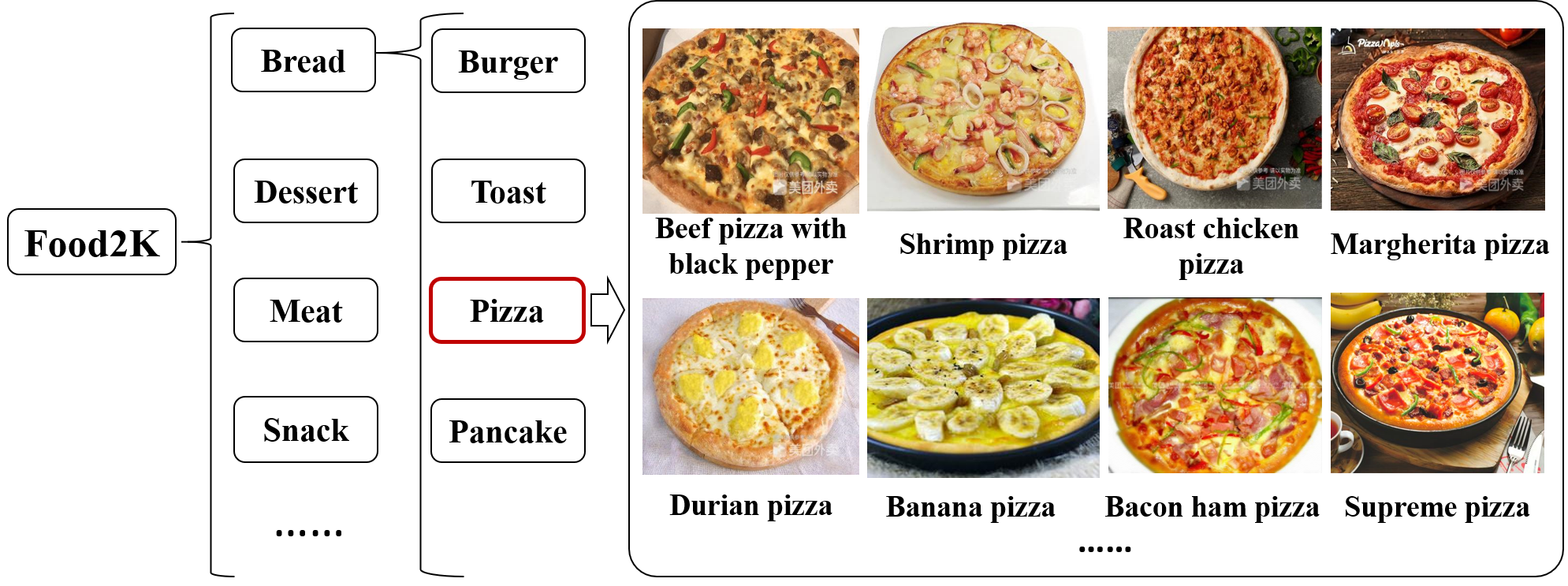}
		\end{minipage}%
	}%
	\centering
	\caption{(a) Various visual appearances for one category and (b) one example with more fine-grained annotation in Food2K.}
	\label{challenges}
\end{figure*}

We collected this dataset from the catering website Meituan\footnote{Meituan is China's leading e-commerce platform for services, especially well known in catering, which is similar to Yelp.}, which  holds a huge number  of  food images including both  eastern and western ones. We first obtained a large-scale raw dataset $S_{o}$  from this website after application and authorization from Meituan. It consists of about seventy million food images uploaded from both catering staffs from take-out online restaurants and users in Meituan. We then processed  $S_{o}$ to  complete the Food2K construction in three phases, namely  food category vocabulary construction, food image collection and labeling.

\textbf{Building a  Vocabulary of Food Categories.}
Considering it is harder to obtain a comprehensive and standard food label list, we adopted a bottom-to-up method to construct the category vocabulary from  noisy and redundant labels associated with images in $S_{o}$. We  first removed special characters and quantifiers, and then aggregated different labels belonging to the same food as one category according to the synonym set of food built from Meituan, leading to one set of  food labels $V_{o}$. However, there still exist many labels corresponding to the same category. We thus carried out the secondary label aggregation. Considering manually judging whether two  food labels belong to the same dish or not is time-consuming and difficult, we conducted the following iterative process: we first selected one food label from $V_{o}$ and manually selected its several representative images from $S_{o}$ (10 in our setting). We  then extracted their deep visual features (Inception-ResNet V2 in our setting), and used their averaged visual features as its visual description  to retrieve  images from $S_{o}$ until we obtained the closest 50 food labels from retrieved labeled images as the candidate set. According to their images and  associated labels, the candidate label set  is  verified to generate the final aggregated label via strict manual verification, and meanwhile  these food labels are removed from $V_{o}$. For example, the following food label set \{Spaghetti with seafood, Italian seafood spaghetti, Angel seafood spaghetti\} is aggregated into the  label `Seafood spaghetti'.  We finally settled on a vocabulary with 2,000 frequently used  categories, covering daily diet with various food types, such as snack, bread and vegetables. Fig.~\ref{Data_Collection_example} (a) shows some categories from the vocabulary.

\textbf{Collecting Food Images. }For each category from the constructed vocabulary, we  aggregated  food images based on its mapping $M$ and then used these aggregated  images  to retrieve more  images from unlabeled ones of $S_{o}$ to enlarge this dataset. Fig.~\ref{Data_Collection_example}~(b) shows an example. The images with the labels from the food category ``Tom yum kung'' can be obtained. An ideal method is to clean up these candidate images via strict manual check. However, it is impractical for huge number of candidate images. Therefore,  we cleaned up these candidate images via  visual retrieval: for each  food category, we similarly chose  representative images (10 in our setting) and obtained their averaged visual features extracted from Inception-ResNet V2 as its visual description. Then we computed the visual similarity between  it and its candidate images, and removed ones with low similarity. To ensure better diversity of food images and lower bias produced from visual retrieval with certain deep network in Food2K, we adopted the relatively lower threshold and for the images within this threshold, different sampling proportions are set according to their different visual similarity range. Different thresholds can be used for different networks to ensure the low bias, and the dependence on certain feature extraction networks is thus small\footnote{Please refer to image samples from some categories via \url{http://123.57.42.89/FoodProject.html}.}. In addition, the original images in the Meituan website are from different sources. For example, the final constructed dataset includes not only food images from merchants, but also self-made food images taken by users themselves. Another way to ensure the diversity is that we aggregated food images based on the label-images mapping. Each category is from various original labels belonging to the same food, making  more diversity of aggregated food images for the fine-grained semantic difference. For example, the corresponding images with the following food label set \{Spaghetti with Seafood, Italian seafood spaghetti, Angel seafood spaghetti\} are attached into the final category label ‘Seafood Spaghetti’.


\textbf{Annotating Food Images.}
After initial food image collection, we conducted exact duplicate removal.  To further build such a high-quality dataset, all the images were finally evaluated by  the professional data annotation and quality inspection team. These annotators are divided into two groups. For each category, we first obtain some groundtruth images  from Wikipedia or encyclopedic items. Based on these reference images, the first group conducted the annotation for each category via removing unqualified images, such as  unclear ones, and ones with food parts occluded. Fig.~\ref{different_forms} illustrated different types of unqualified images. Note that all the images in Food2K have a single dish label, since  our primary  goal is to provide the basic large-scale food recognition benchmark to support more complex food-relevant vision tasks. The second group then carried out the inspection for each  image from this category to ensure that the rate of data qualification reaches 99\%. Otherwise, the images belonging to this category should be re-annotated. Through the iterative annotation and multiple inspection, the high-quality Food2K dataset is finally built with 2,000 classes and 1,036,564 images.

\subsection{Dataset Statistics and Analysis}
\label{DatasetStatistics}
Considering Food2K contains both western and eastern food,  with the help of food experts, we built a unified food ontology  by combining  and adapting existing food classification systems from both western~\cite{nestle2013food,Singla-FnFC-MM2016} and eastern ones~\cite{food_safety}. It is a hierarchical structure: there are 12 super-classes (like “Bread” and “Meat”), and there are some sub-classes for each super-class (like ``Beef” and ``Pork” in ``Meat”). Each type contains lots of dishes (like ``Curry Beef Brisket” and ``Fillet steak” in ``Beef”), where each dish generally contains several ingredients. Note that  some groups are very broad in the ontology, like ``Bread" and ``Meat", and are  thus divided into more detailed types. As shown in Fig.~\ref{Hierarchy-tree_ontology}, 2,000 dish categories from Food2K covers all the 12 groups with both eastern food (e.g., Noodles and Sushi) and western food (e.g., Pasta and Dessert). Food2K contains 1710 eastern food categories  and 290 western ones. Fig.~\ref{distribution2} shows the number of images per category with decreasing order. The number of images per category is in the range $[153,1999]$, showing a larger class imbalance compared with existing food datasets. Some samples\footnote{Please refer to the full category list and  samples from all  categories via \url{http://123.57.42.89/FoodProject.html}.} are also showed in Fig.~\ref{distribution2}.

Fig.~\ref{Data_Collection_ComP} illustrates the scale of Food2K, compared to existing food recognition datasets. As shown in  Fig.~\ref{Data_Collection_ComP}~(a), the size of images from Food2K  bypasses the size of the previous largest one by one order of magnitude. Fig.~\ref{Data_Collection_ComP}~(b) provided an overview of distributions of dishes on categories and compared to  typical datasets, including ETH Food-101 (western food), Vireo Food-172 (eastern food) and ISIA Food-500 (Misc. food). We can see that for each category, the number of dishes from Food2K is larger than existing datasets. Furthermore, some categories exist in Food2K, but missing or very few in existing ones, such as Barbecue. Further statistics show the overlapped categories between existing datasets (Food-101, Vireo Food-172 and ISIA Food-500) and Food2K are only 13, 75 and 40, respectively, and duplicate detection shows that there are no repeated images for each overlapped category. The probable reason is that images from Meituan are taken by catering staffs and users in Meituan, and are not allowed crawled without authorization.
\begin{figure*}[htb]
	\centering
	\includegraphics[width=0.9\textwidth]{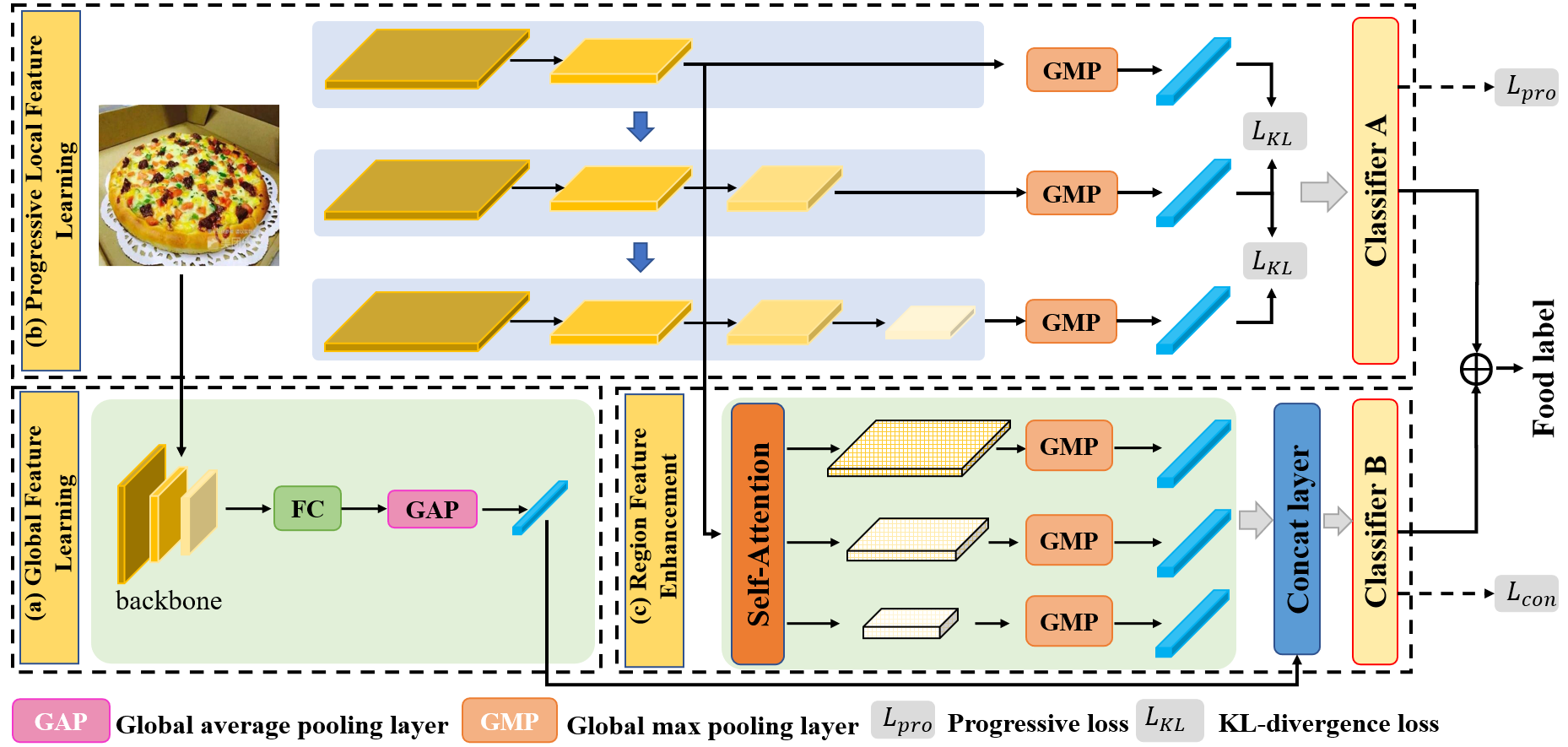}
	\caption{The framework of PRENet. (a) Global Feature Learning branch, which learns the global superclass features. (b) Progressive Local Feature Learning branch, which capture complementary multi-scale local features through progressive training strategy. (c) Region Feature Enhancement branch, which incorporate contexts into local features through self-attention. Note that the predicted scores by classifier A in (a) and B in (b) are combined for the final prediction.}
	\label{method1}
\end{figure*}
\label{Our_method}
Besides  more comprehensive  coverage of dishes and larger quantity of images, Food2K has  the following characteristics: First, Food2K covers more diverse  visual appearance and patterns. Fig.~\ref{challenges}~(a) shows some examples for each category. Different ingredients and their combination, different accessories, different arrangements all lead to the visual difference for the same category. For example, the fresh fruit salad appears different visual appearances for its  mixture of different fruit ingredients, while the large visual difference of “Rib eye steak" comes from its different accessories. Such unique characteristics from food leads to higher intra-class difference, making large-scale food recognition difficult. Second, Food2K contains more fine-grained annotation. Based on the constructed ontology of Food2K, the level of category annotation in Food2K is  more fine-grained compared with other food datasets. Take Pizza as an example, some classic datasets such as Food-101 has only the pizza class. In contrast, as shown in Fig.~\ref{challenges}~(b), the pizza class in Food2K is further divided into more types. Their subtle visual differences among different pizza images are mainly caused by their unique ingredients or the same ingredient with different granularity, also leading to more difficulty in recognition. All of these factors enable Food2K one new challenging large-scale food recognition benchmark.

\section{Our Method}
In this section, we present the proposed  Progressive Region Enhancement Network (PRENet) in Fig.~\ref{method1}. PRENet  mainly  consists of progressive local feature learning and region feature enhancement. The former  mainly adopts the progressive training strategy to  learn complementary multi-scale finer local features, like different ingredient-relevant information. The region feature enhancement uses self-attention to incorporate richer contexts with multiple scales into local features to enhance the local feature representation. Then we fuse enhanced local features and global ones from global feature learning into the unified one via the concat layer. During training, after  progressively training the networks from different stages, we  then train the whole network with the concat part, and further introduce the KL-divergence to  increase  the  difference between stages for capturing more detailed features. For the inference, considering  the  complementary output from  each stage and the concatenated features, we combine the prediction results from them for final food classification.

\subsection{Global-Local Feature Learning}
Although food image recognition is one fine-grained visual recognition task, the food images under different sup-classes have obvious discriminative visual differences, and thus can be better recognized by global representation. Those in different sub-classes under the same super-class have high inter-class similarity, just as shown in Fig.~\ref{challenges} (b), and we thus should pay more attention to more fine-grained local features.

Therefore, we extract and fuse both global representation and their subtle visual differences. We use two sub-networks to extract the global and local features of food images respectively. These two sub-networks can be two separate networks. However, they share most of the layers of the same backbone network in our network for efficiency.\\
\textbf{Global Feature Learning} Inspired by~\cite{zhang2019learning}, Based on existing network (e.g., ResNet), for the output $f^{g}$ of the last convolutional layer, we use Global Average Pooling~(GAP) to extract the global feature $f_{Glo}$ :
\begin{equation}
	\begin{aligned}
		f_{Glo}=\mathbf{GAP}(f^{g})\\
	\end{aligned}
\end{equation}
\\
\textbf{Progressive Local Feature Learning} Local features sub-network aims to learn discriminative fine-grained features of food. Due to the diverse ingredients and cooking style, the discriminative parts of the food image are multi-scale and irregular. As the first contribution, we adopt the progressive training strategy to solve this problem. In this strategy, we train the low stages first which have the small receptive field, then zoom out a larger field surrounding this local region, and finish when we reach the whole image. This training strategy will force our model to extract finer discriminative local features, such as ingredient-relevant ones. After this process, we extract features from different layers to obtain multi-scale feature representations. Specifically, for the output $f^i$ of each stage from the local feature sub-network, we use a convolutional block and a Global Maximum Pooling~(GMP) layer to get their local feature vectors $f_{Loc}^i$: 
\begin{equation}
	\begin{aligned}
		f_{Loc}^i=\mathbf{GMP}(f^i)
	\end{aligned}
\end{equation}
where $f^i$ denotes the output from the $i^{th}$ stage of the network.


Therefore, this strategy first learns more stable finegrained information in shallower layers and gradually shifts attention to learning coarse-grained information in deeper layers as training progresses. Specifically, it can extract discriminative local fine-grained features such as ingredients when features with different granularities are sent to the network. However, simply using progressive training strategy will not get diverse fine-grained features, because the multi-scale information learned via progressive training may focus on the similar region. As the second contribution, we optimize the KL divergence between features from different stages to increase the difference between them to solve the problem. By maximizing the KL divergence between features from different stages, we force multi-scale features to focus on different areas in different stages, which can help capture as many details as possible.

Particularly, we divide the training process into $S$ steps, and train the first $U-S+i$ stages at step i, where $U$ is the total number of stages of the network. Because we will concatenate all the global and local features in the final stage (also called concat stage), thus the total number of steps is $S +1$. In our method, for the output $f_{Loc}^i$ from the $i^{th}$ stages of the network to be trained in each step of progressive training, $F_i$ is used to process the output features. $F_i$ consists of a convolutional layer, a batch norm layer and a ReLU unit. Then we can get local feature representation $c^i=F_i(f_{Loc}^i)$.


Furthermore, for multiple local features from different stages, we utilize KL-divergence to increase the difference between stages, which can help capture as many details as possible. Under the progressive training strategy, the visual features extracted from different stages are projected as specific probability distribution which represent the visual semantic information. However, the iterative optimization may result in different probability distributions converging to the same distribution, which impairs the ability to extract diverse features. KL-divergence can measure the similarity between two distributions. By maximizing the KL-divergence, the convergence can be suppressed and more fine-grained visual features can be extracted for recognition. The KL-divergence is calculated over global features for every adjacent two outputs in each batch, where the reduction of KL-divergence is batchmean.
\begin{equation}
	\begin{aligned}
		&L_{KL}(y_i,y_j)=\sum_{i=1}^{U}\sum_{j=U-i}^{U} y_i log(\frac{y_i}{y_j})\\
	\end{aligned}
\end{equation}
where $y_i$ and $y_j$ are the output distribution from different stages.  

\subsection{Region Feature Enhancement}
Different from general fine-grained tasks, food images do not have fixed semantic information~\cite{Min2019A}. Most of existing food recognition methods~\cite{Martinel-WSR-WACV2018,Min-IGCMAN-ACMMM2019} mine these discriminative features directly, ignoring  the relationship between local features. 
Therefore, we adopt a self-attention mechanism to capture the relationship between different local features. This strategy aims to capture the co-occurring food features in the feature map. It is revised non-local interaction~\cite{wang2018non} within the same level feature map, and the output feature map has the same scale as its input. Specifically, we first extract the local feature representation $f_{Loc}$ of 
the last $S$ stages, and then obtains the enhanced features via self-attention as follows:
\begin{equation}
	\begin{aligned}
		&q^{(i)}=\mathbf{Conv}(f_{Loc}^{(i)})\\
		&k^{(j)},v^{(j)}=\mathbf{Conv}(f_{Loc}^{(j)})\\
		&S_{i,j}=\mathbf{Softmax}(q^{(i)}k^{{(j)}^T}/\sqrt{d_{k^{(j)}}})\\
		&\widehat{f_{IJ}}={S_{i,j}} v^{(j)}\\
	\end{aligned}
\end{equation}
where $f_{Loc}^{(i)}$ and $f_{Loc}^{(j)}$ are the $i^{th}$ and $j^{th}$ feature positions in $f_{Loc}$. $q^{(i)}$ is the $i^{th}$ query and $k^{(j)},v^{(j)}$ are the $j^{th}$ key/value pair, $d_{k^{(j)}}$ is the dimension of the $k^{(j)}$ and $S_{i,j}$ means the similarity between $q^{(i)}$ and  $k^{(j)}$. Thus we can get the enhanced feature $\widehat{f_{IJ}}$. Finally, we concat these enhanced feature maps of the same size, and use convolutional layers to convert them into the same dimension. Finally, we can get $S$ local features $\widehat{f^i_{Loc}}$ after it. By this strategy, our model enables features to interact across space and scales.

During training, after obtaining  global features and  local features, we combine them as the final representation $f'$ in the concat stage:
\begin{equation}
	\begin{aligned}
		&f'={\rm{Concat}}(f_{Glo},\widehat{f^{U-S+1}_{Loc}},……,\widehat{f^U_{Loc}})
	\end{aligned}
\end{equation}

\subsection{Optimization and Inference}
During the optimization, we use an iterative procedure to update the parameters of the network. First, during progressive learning stage, we utilize the cross entropy loss $L_{pro}^i$ from $S$ stages to back propagate for updating the parameters of the corresponding network. Note that all the parameters in the current stage will be optimized, even they have been updated in the previous stages. Then, during the concat stage, we utilize another  loss function to update the parameters of the whole network. Our network is trained  in an end-to-end way.

During the progressive training, for the output from each stages, we can utilize classifiers $\mathbf{C}_{A}^i$ to predict the corresponding probability distribution as:
\begin{equation}
	\begin{aligned}
		y_i=\mathbf{C}_{A}^i(c^i)
	\end{aligned}
\end{equation}
where classifiers $\mathbf{C}_{A}^i$ consists of two fully-connected stage with batchnorm and Elu non-linearity, corresponding to the $i^{th}$ stage. $y_i$ means the prediction from the corresponding stages, and i ranges from $U-S+1$ to $U$. Then, we adopt the following cross entropy loss:
\begin{equation}
	\begin{aligned}
		L_{pro}^{i}=-\sum_{x^m\in{M}}{y_i^m lny_i^m}
	\end{aligned}
\end{equation}
where $M$ is a set of training data, $x^m$ means the $m^{th}$ sample and $y_i^m$ is the corresponding category label. 

For the unified representation, the prediction from the final Concat stage can be obtained by a classifier $\mathbf{C}_{B}$ as:
\begin{equation}
	\begin{aligned}
		&y'=\mathbf{C}_{B}(f')\\
	\end{aligned}
\end{equation}

we use another cross entropy loss as:
\begin{equation}
	\begin{aligned}
		&L_{con}=-\sum_{i=1}^{N}{{y'}_i \times ln {{y'}_i}}\\
	\end{aligned}
\end{equation}
where $y'$ is the predicted label, $N$ is the total number of the training samples . In our method, the $L_{con}$ is kept for all runs and number of steps is $S +1$. 
Furthermore, we introduce another loss of KL-divergence to increase  the  difference between stages for capturing more detailed features, resulting in the final loss function for the whole network:
\begin{equation}
	\begin{aligned}
		&L=\alpha L_{con}+\beta L_{KL}\\
	\end{aligned}
\label{totalLoss}
\end{equation}
where $\alpha$ and $\beta$ are balance parameters. 

During the inference step, considering the prediction from different stages and the fused stage is complementary,  we can combine all outputs together to improve the recognition performance. Particularly,  we add the scores across all stages (including the final score) with equal weights to predict the output class:
\begin{equation}
	\begin{aligned}
		&Y={\rm{Sum}}(y',y_{U-S+1},...,y_{U})\\
	\end{aligned}
\end{equation}
where ${\rm{Sum}}()$ denotes the weighted sum with the equal weights


\section{Experiment}
We first begin with the extensive evaluation of our proposed method for food  recognition on Food2K.  Then, we study the generalization ability of Food2K on five food-relevant  tasks, including food recognition, food image retrieval, cross-modal recipe retrieval, food detection and food segmentation. Finally, we  discuss  potential research problems and techniques that could be investigated based on Food2K.

\subsection{Recognition on Food2K}
\label{Exper-BR}
\subsubsection{Implementation Details}
Food2K is divided into 60\%, 10\% and 30\% images for training, validation and test set, respectively. Top-1  (Top-1 Acc.) and Top-5  classification accuracy  (Top-5 Acc.) are adopted as evaluation metrics. All the neural network models are implemented using the PyTorch framework. 

We  compare various methods, including   deep  networks~(e.g., DenseNet161~\cite{huang2017densely} and SENet~\cite{Jie2017Squeeze}), recently proposed fine-grained recognition methods~(e.g., MOMN~\cite{2020Multi} and PMG~\cite{du2020fine}) and  food recognition methods~(e.g., PAR-Net~\cite{Qiu-MDFR-BMVC2019}).  For the deep networks, we train all the networks with parameters initialized from  ImageNet pre-trained weights  with a learning rate of $10^{-2}$, and divided by 10 after 30 epoches. All the networks are optimized using the stochastic gradient descent with a momentum of 0.9, and weight decay of $10^{-4}$. Training and testing are performed with an image size of $224 \times 224$, excepting  Inception family uses the size of $299 \times 299$. For fine-grained methods and food recognition methods, all the presented experiments follow the same settings in the mentioned papers. Similar to fine-grained recognition methods, in our method, the input images are resized to a fixed size of 550 $\times$ 550 and randomly cropped into 448 $\times$ 448, and random horizontal flip and color jitter are applied for data augmentation when we train the model.  The test images are resized to a fixed size of 550 $\times$  550 and cropped from center into 448 $\times$  448. The initial learning rate is $10^{-3}$ and multiplied by 0.9 after 2 epoches. Note that for fair comparison in fine-grained methods, we list results with the same backbone mentioned in their papers. For our method, the number of learning stages $U=3$, which corresponds to the last $3$ layers of the network. The number of steps $S=3$, and the balance parameters of the total loss $\alpha=0.8$, $\beta=0.2$.
\begin{table}[!t]
	\small
	\caption{Comparison of our approach (PRENet) to baselines on Food2K~(\%).}
	\begin{center}
		\begin{tabular}{ccc}
			\hline
			{\textbf{Methods }} &Top-1 Acc. & Top-5 Acc.\\
			\hline		
			VGG16~\cite{Szegedy-GDC-CVPR2015}& 78.96&95.26   \\
			Inception v4~\cite{Szegedy-Inception-v4-AAAI2017}& 82.02 &96.45 \\
			ResNet50~\cite{He-DRL-CVPR2016}&80.79&95.74 \\
			ResNet101~\cite{He-DRL-CVPR2016}&81.28&95.99 \\
			ResNet152~\cite{He-DRL-CVPR2016}& 81.95&96.57 \\	
			Inception-ResNet-v2~\cite{Szegedy-Inception-v4-AAAI2017} & 82.07&96.74 \\
			WRN-50-2-bottleneck~\cite{Zagoruyko-WRN-BMVC2016}& 81.94&96.19\\	
			DenseNet161~\cite{huang2017densely}&  81.87&96.53 \\
			SE-ResNeXt101\_32x4d~\cite{Jie2017Squeeze} &80.81&95.61 \\
			Inception v4~\cite{Szegedy-Inception-v4-AAAI2017}&82.02&96.45\\	
			SENet154~\cite{Jie2017Squeeze}&83.62 &97.22 \\	
			\hline
			NTS-NET(ResNet50)~\cite{Yang-L2Nav-ECCV2018}& 81.24&94.94 \\
			HBP(ResNet50)~\cite{Yu_2018_ECCV}& 77.56&92.87\\
			DCL(ResNet50)~\cite{Chen_2019_CVPR}& 82.15&96.72  \\
			WS-DAN(ResNet50)~\cite{Hu_2019_CVPR}&81.37&96.27\\
			Inception v4 @ 448px~\cite{Kornblith2018Do}&  82.46&97.17\\
			MOMN(ResNet50)~\cite{2020Multi}&80.84&96.02\\
			PMG(ResNet50)~\cite{du2020fine}&81.29&96.12\\		
			\hline
			DLA~\cite{Yu-DLA-CVPR2018}&80.14&96.37\\
			PAR-Net(ResNet101)~\cite{Qiu-MDFR-BMVC2019}&80.93 &96.61\\
			\hline
			PRENet(ResNet50)&{83.03}&{97.21}\\
			PRENet(ResNet101)&\textbf{83.75}&\textbf{97.33}\\
			\hline
		\end{tabular}
	\end{center}
	\label{baselines}
\end{table}

\begin{figure}
	\centering
	\includegraphics[width=0.5\textwidth]{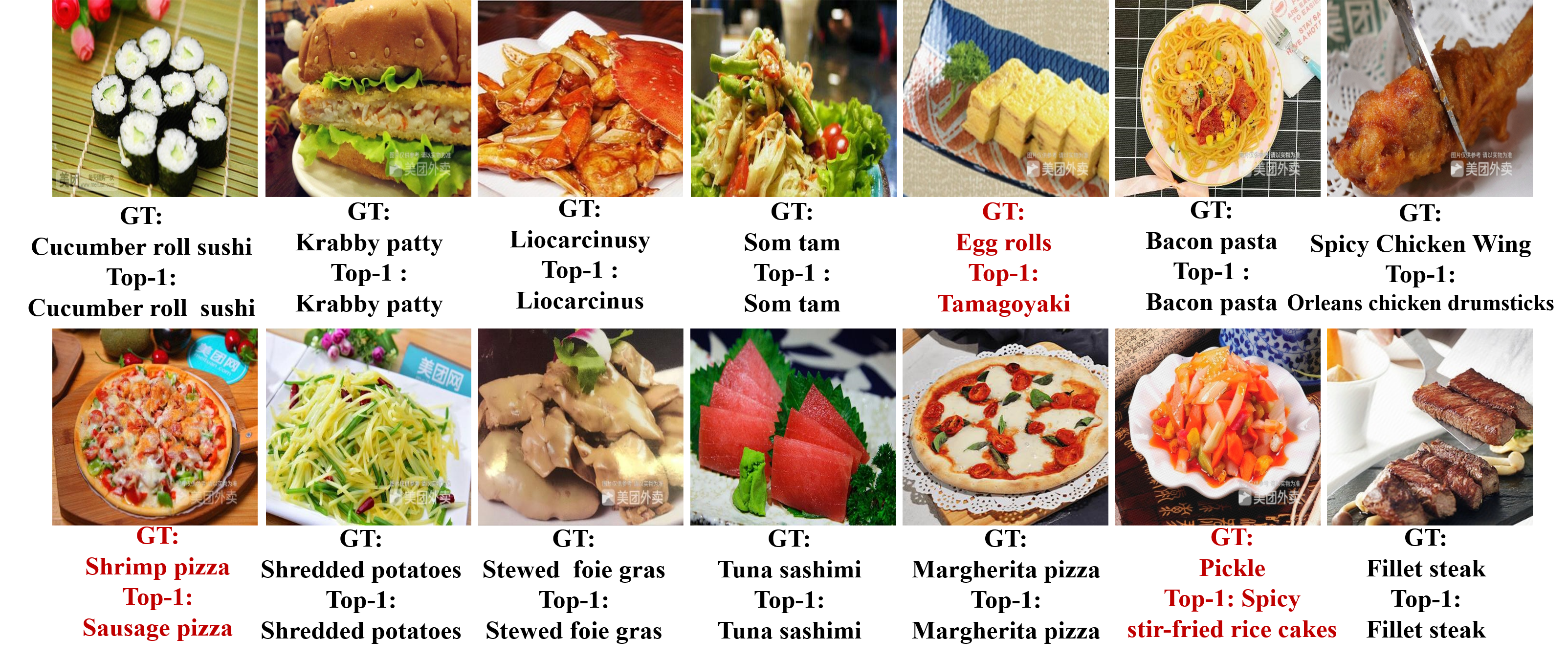}
	\caption{Some classification results on Food2K. GT means the ground truth. The dishes
	with red color are not correctly classified in Top-1 results.}
	\label{recognition_results}
\end{figure}

\subsubsection{Comparisons with State-of-the-Art Methods}
Table~\ref{baselines} shows the performance comparison of different methods on Food2K. We can see that  the recognition performance has consistent improvement when more advanced networks are adopted, indicating  deeper and more advanced networks for further improvement. When using the same backbone, we further observe that most  fine-grained methods perform better than baseline networks, such as DCL~\cite{Chen_2019_CVPR}. However, the performance trend on Food2K is not consistent with existing fine-grained datasets, such as birds and cars. For example, recently proposed PMG~\cite{du2020fine} does not achieve the desired performance on Food2K compared with existing fine-grained datasets. The probable reason is that PMG does not consider the relations between  local features and it also does not take the global representation into consideration during feature learning. 
There are  also some fine-grained methods, which  perform even worse than backbones, such as HBP~\cite{Yu_2018_ECCV}. The probable reason is that HBP adopts the bilinear pooling to capture the inter-layer part feature relations and extracts the co-occurring features to integrate a unified representation for classification. Therefore, the model may focus more on the same and common semantic parts, such as the bird’s mouth.  However, many food categories have  non-rigid structures, and they do not have fixed semantic information. These experimental results  demonstrate that directly adopting existing fine-grained methods do not necessarily achieve the optimal performance for large-scale food recognition, indicating that we should design food-oriented network  for further performance improvement. 

The performance of  food recognition methods such as PAR-Net  also does not achieve reasonable results. We speculate that these food recognition methods are more suitable for existing small or medium-scale food datasets, such as ETH Food-101 and VireoFood-172. From Table~\ref{baselines}, we can see that our method achieves competitive result on Food2K. Our method outperforms the backbone~(ResNet50) by  2.24 \% and 1.47 \% in Top-1  and Top-5 classification accuracy, respectively, and it also outperforms PMG~\cite{du2020fine} by 1.74\% in Top-1 classification accuracy, even PMG adopts various granularities to learn  local features. This verifies the advantage of  combining  the progressive training strategy and self-attention to enhance local feature representation.

We further show  responses to some predicted examples  by PRENet~(ResNet50)  in Fig. \ref{recognition_results}. We observe that  there are still some wrongly predicted results, and they  are mainly from  more fine-grained confusion, e.g., ``Shrimp pizza" and ``Sausage pizza", ``Pickle" and ``Spicy stir-fried rice cakes",  revealing opportunities for algorithmic improvements to handle these  challenges.

\subsubsection{Ablation Study on PRENet}
We conduct various ablation studies to understand the effectiveness of  our method from different aspects, where  ResNet50 is used as the backbone network. 
\begin{figure*}[htb]
	\centering
	\includegraphics[width=1.0\textwidth]{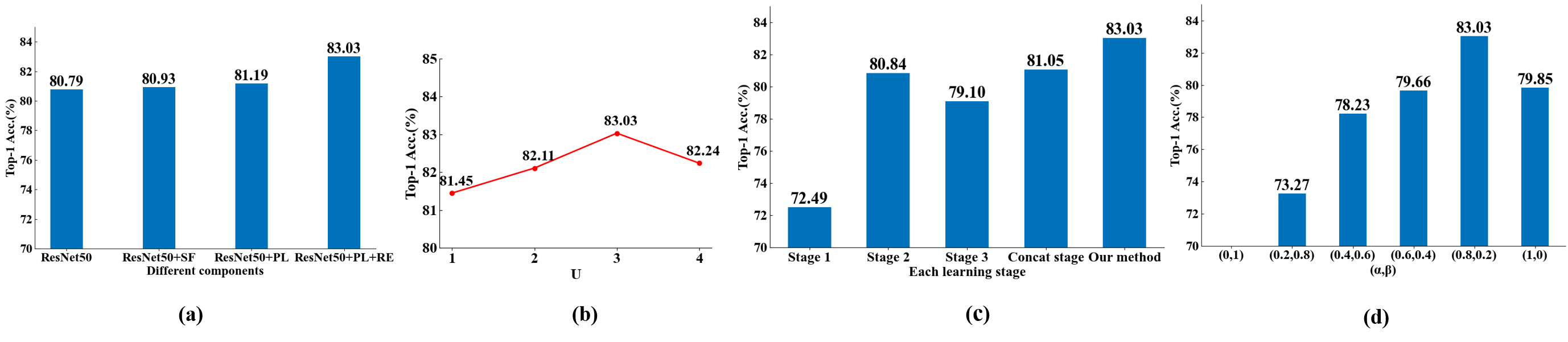}
	\caption{Ablation study of PRENet on Food2K: (a)~Different components. (b)~Different number of learning stages $K$. (c)~Each learning stage. (d)~Different balance parameters ($\alpha$,$\beta$).}
	\label{concat5}
\end{figure*}\\
\textbf{Effect of Different Components}  We study the effect of different components in our method, including  progressive learning strategy~(PL), region feature enhancement~(RE) and their combination. For comparison, we introduce another baseline Simple Fusion~(SF), which simply uses concatenated features from the last three layers without PL and RE for recognition. As shown in Fig.~\ref{concat5}~(a), we can see that: (1) The introduction of PL   brings the recognition performance gain; (2) The combination of PL and RE gives further performance boost, which  shows that our method is effective in learning  and enhancing local features via both PL and RE.\\
\textbf{The Number of Learning Stages $U$} We study the effect of our method when changing the number of learning stages $U$. The results are reported in Fig.~\ref{concat5}~(b). It is clear that  increasing $U$ boosts the model performance. Our model achieves 81.45\%, 82.11\% and 83.03\% Top-1 classification accuracy from $U=1$ to $3$ on Food2K. The classification performance of our model  receives consecutive gains by 0.66\% and 0.92\%. However, we  notice the accuracy starts to decrease when $U=4$. The possible reason is that low stage layers  mainly focus on class-irrelevant features. With deep  progressive training,  too many stages probably force the model to find class-relevant information and  introduce the noise to this class, which may cause the model to generalize poorly on the test set and the overall performance probably decreases.  \\
\textbf{Effect of Different Learning Stages}  To better verify the contribution of each learning stage and the final concat stage, we also test the accuracy by  using the predication from  each stage separately. The results are reported in Fig.~\ref{concat5}~(c). We observe that the Concat learning stage achieves the best performance for Top-1 classification accuracy.  It can reveal that our method  captures and fuses  complementary information from different stages, and achieves the best recognition performance.
\begin{figure}[htbp]
	\centering
	\includegraphics[width=0.48\textwidth]{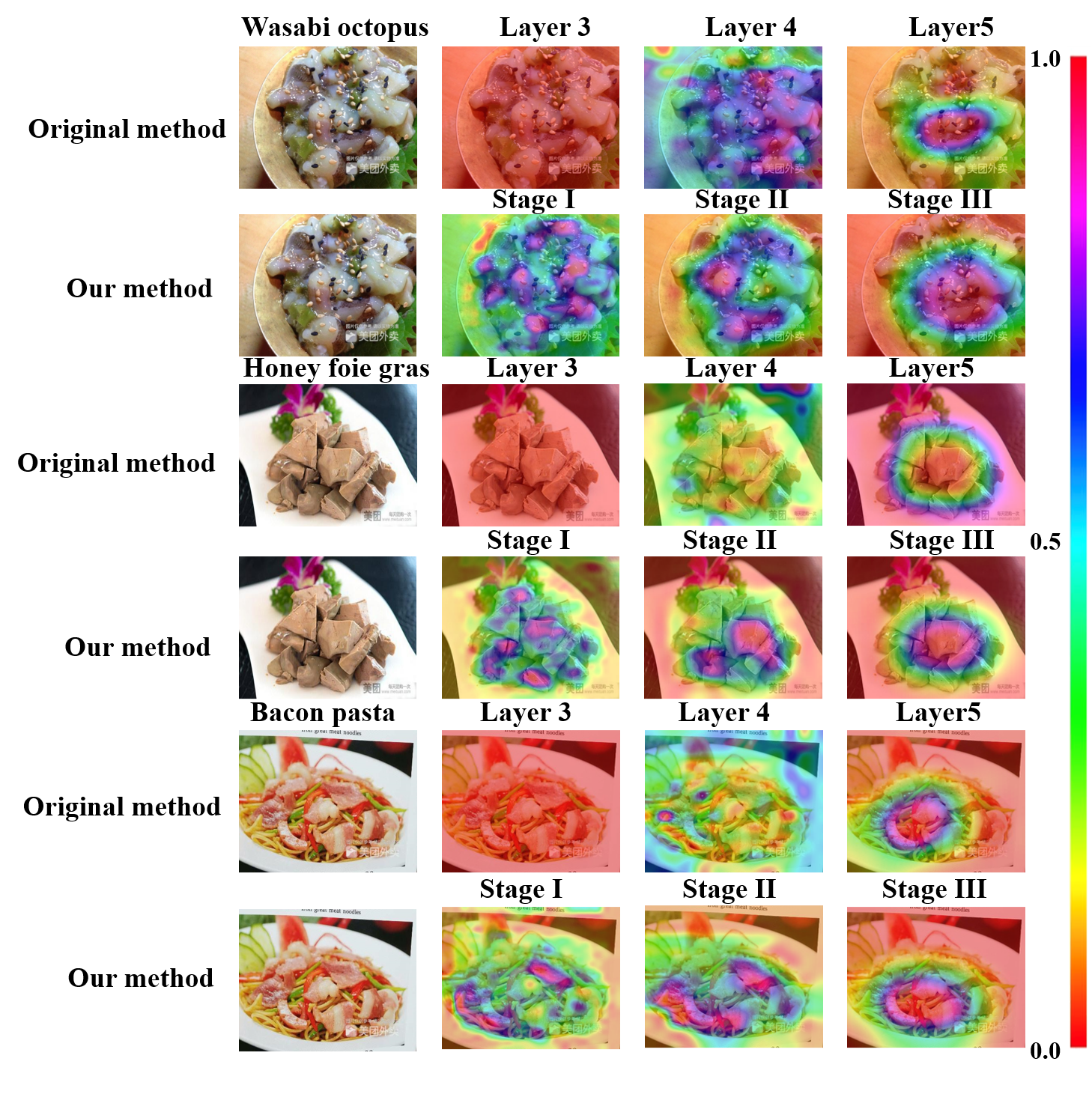}
	\caption{Visualization results of proposed  progressive learning on some samples from Food2K. The original method means we use  feature maps from last three layers of the network without PL for visualization.}
	\label{visualization}
\end{figure}\\
\textbf{Balance Parameters $\alpha$  and $\beta$} We study the influence of  two balance parameters  $\alpha$  and $\beta$  in the total loss Eq.~\ref{totalLoss}. As shown in Fig.~\ref{concat5}~(d),  when $\alpha=0$ and $\beta =1$, the total loss is only optimized by the KL divergence, and the model can not converge. With $\alpha$ increasing, our model performs better until it  reaches a tipping point when $\alpha=0.8$ and $\beta =0.2$. When $\alpha=1$,  our model is only optimized by the cross entropy loss, and the Top-1 classification accuracy is reduced by 3.18\%, which  proves that introducing the KL divergence can obtain  better performance. This is because KL divergence can force multi-scale features to focus on different areas, and thus  helps capture as many details as possible.

\subsubsection{Visualization Analysis}


\begin{table}[htbp]
	\caption{Performance comparison on ETH Food-101 (\%).}
	\begin{center}
		\begin{tabular}{ccc}
			\hline
			\textbf{Method} &Top-1 Acc. & Top-5 Acc.\\
			\hline
			AlexNet-CNN~\cite{Bossard-Food101-ECCV2014}&56.40&-\\
			SELC~\cite{Martinel2016A}&55.89&-\\
			ResNet-152+SVM-RBF~\cite{mcallister2018combining}&64.98&-\\
			DCNN-FOOD(AlexNet)~\cite{Yanai-FIRDCNN-ICME2015}&70.41&-\\
			LMBM(GoogLeNet)~\cite{Wu-LMBM-MM2016}&72.11&- \\
			EnsembleNet~\cite{Pandey2017FoodNet}&72.12&91.61\\
			GoogLeNet~\cite{Ao2015Adapting}&78.11&- \\
			DeepFOOD(GoogLeNet)~\cite{Liu2016DeepFood}&77.40&93.70\\
			ILSVRC~\cite{Bola2017Simultaneous}&79.20&94.11\\
			WARN(WRN-50)~\cite{Rodriguez-PATA-TMM2019}&85.50&- \\
			CNNs Fusion(I$_{2}$)~\cite{Aguilar2017Food}&86.71&-\\
			Inception V3~\cite{Hassanne-FIRDCN-MM2016}&88.28&96.88\\
			SENet-154~\cite{Jie2017Squeeze}& 88.62 &97.57 \\
			SOTA\cite{Kornblith2018Do}&90.00&- \\
		    NTS-NET(ResNet50)~\cite{Yang-L2Nav-ECCV2018}&89.40&97.80\\
			HBP(ResNet50)~\cite{Yu_2018_ECCV}&86.23&97.13\\
			DCL(ResNet50)~\cite{Chen_2019_CVPR}&88.90&97.80 \\
			PMG(ResNet50)~\cite{du2020fine}&86.93&97.21\\
			WS-DAN(Inceptionv3)~\cite{Hu_2019_CVPR}&88.90&98.11 \\    
			Inception v4 @ 448px\cite{Kornblith2018Do}&90.00&- \\
			DLA\cite{Yu-DLA-CVPR2018}&90.00&- \\
			PAR-Net(ResNet101)~\cite{Qiu-MDFR-BMVC2019}&89.30&-\\
			WARN(WRN)~\cite{martinel2018wide}&85.50&-\\
			IG-CMAN(SENet154) ~\cite{Min-IGCMAN-ACMMM2019}&90.37&98.42\\
			MSMVFA(SENet154)~\cite{Min-MSMVFA-TIP2019}&90.59&98.25\\
			SGLANet(SENet154) ~\cite{Min-ISIA-500-MM2020}&89.69&98.01\\
			\hline
            PRENet~(ResNet50)&89.91&98.04\\
            PRENet~(ResNet50+Pretrained)&89.99&97.87\\
            PRENet~(ResNet101)&90.32&97.84\\
            PRENet~(ResNet101+Pretrained)&90.69&98.65\\
			PRENet~(SENet154)&90.74&98.48\\
	  	PRENet~(SENet154+Pretrained)&\textbf{91.13}&\textbf{98.71}\\
			\hline
		\end{tabular}
	\end{center}
	\label{101recognition_performance}
\end{table}

To gain further insight into our method, we further conduct visualization analysis via Grad-CAM~\cite{Selvaraju2017Grad}. We visualize the output feature maps from different learning stages and compare our method with the original one in Fig.~\ref{visualization}. The original method means we use  feature maps from last three layers of the network without PL and RE for visualization. For these typical examples, the attentional regions are expanded with the stages going ahead, and  more discriminative and detailed parts have been included. Moreover, our method can capture various local features in different stages for the introduction of KL-divergence. Take ``Wasabi octopus" for example (the second row in Fig.~\ref{visualization}), the baseline only  obtains limited information and different feature maps tend to focus on the similar part. In contrast, for our method, the Stage I  pays more attention to the ``vegetable leaf", while the Stage II mainly focuses on the ``octopus". For Stage III, the overall characteristics of this food can be captured, and thus both global and local  features are utilized for recognition.

\subsubsection{Recognition Performance on Other Datasets}
Considering ETH Food-101 has been one standard benchmark in the computer vision, we also conduct the evaluation on this dataset to verify the effectiveness of the proposed method. As shown in Table~\ref{101recognition_performance}, we can observe that when adopting the same  setting, our method achieves the best performance compared with  existing methods for Top-1 classification accuracy. For example,  our method outperforms  typical fine-grained methods with  ResNet50 as the backbone. Compared with  existing food recognition methods, such as MSMVFA~\cite{Min-MSMVFA-TIP2019}, our method also obtains the highest accuracy 90.74\% for Top-1 classification accuracy. When we use the trained backbone model from Food2K, namely PRENet~(SENet154+Pretrained), there is further performance improvement.

\subsection{Generalization Ability of Food2K}
In this section, we conduct comprehensive evaluation on the generalization ability of Food2K in various vision and multimodal tasks, including food  recognition, food image retrieval, cross-modal recipe retrieval, food detection and segmentation.
\subsubsection{Food Recognition}
We assess the generalization of  models learned using Food2K to ETH Food-101. In addition, we also conduct the evaluation on another two datasets, namely Vireo Food-172  and ISIA Food-500 from the multimedia field. All of  presented experiments follow the same training-test-splitting in the mentioned papers. Representative methods including baseline networks, fine-grained recognition  and food recognition methods are used for evaluation. For each method, there are two settings. Take VGG16 for example: VGG16 denotes we use the target dataset to fine-tune the ImageNet pre-trained network, while VGG16 + Fine-tuned on Food2K denotes we first use Food2K to fine-tune the ImageNet pre-trained network, then fine-tune it on the target dataset for evaluation.

We report experimental results  in Table~\ref{Transfer-recognition}. From Table~\ref{Transfer-recognition}, we can see that all the transferred features are better  than training on target datasets alone. For example, the network fine-tuned on Food2K can improve  Top-1  classification accuracy by 7.33\%, 2.58\%, 2.31\%, 2.61\% and 2.74\%  for VGG16, ResNet152, Inception V3, DenseNet161 and SENet154, respectively on Vireo Food-172. These results show that features learned on Food2K  generalize well on food recognition. The average performance improvement on several common popular neural networks including VGG16, ResNet152, Inception V3, DenseNet161 and SENet154 is 1.68\%, 3.51\% and 3.41\% in Top-1 classification accuracy for ETH Food-101, Vireo Food-172 and ISIA Food-500, respectively, indicating that higher performance gain  is from Vireo Food-172 and ISIA Food-500 while the lowest performance gain is from ETH Food-101. The probable reason is as follows: both ISIA Food-500 and Food2K are Misc. (including both eastern and western cuisines). Therefore, their domain gap is relatively small. In addition, there is a larger proportion of eastern food categories in Food2K and Vireo Food-172 consists of eastern cuisines, also resulting in the smaller domain gap between Food2K and Vireo Food-172. For more complicated networks, we also observe higher gain in Vireo Food-172. Take PAR-Net as an example, the performance improvement is 0.63\%, 1.32\% and 0.74\% in Top-1 classification accuracy for ETH Food-101, Vireo Food-172 and ISIA Food-500. 

We further compare the performance of transfer learning from different food datasets, and give discussions on the recognition performance with trained models based on different ratios of Food2K. Details are in supplementary materials.


\begin{table*}[htbp]
	\small
	\caption{Results of transferring visual representations learned on Food2K to three datasets (\%).}
	\begin{center}
		\begin{tabular}{lcccccc}
			\hline
			\multirow{2}*{\textbf{Models}} &
			\multicolumn{2}{c}{ETH Food-101}&\multicolumn{2}{c}{Vireo Food-172}&\multicolumn{2}{c}{ISIA Food-500}\\
			\cline{2-7}&Top-1 Acc. & Top-5 Acc. & Top-1 Acc. & Top-5 Acc. & Top-1 Acc. & Top-5 Acc.\\			 
			\hline
			VGG16~\cite{Szegedy-GDC-CVPR2015}& 79.02 &93.78  &80.41 &94.59 & 55.22&82.77\\			
			+Fine-tuned on Food2K& \textbf{80.68}&\textbf{94.45}& \textbf{87.74}&\textbf{97.57} &\textbf{56.59} &\textbf{83.65}\\
			\hline
			ResNet50&84.50&96.18&87.32&97.01&56.75&82.80\\
			+Fine-tuned on Food2K&\textbf{85.89}&\textbf{96.66}&\textbf{89.46}&\textbf{97.84}&\textbf{57.97}&\textbf{83.40}\\
			\hline
			ResNet152~\cite{He-DRL-CVPR2016}& 86.61 &96.95  & 86.86&97.11 & 57.03&83.80\\			
			+Fine-tuned on Food2K& \textbf{87.58}&\textbf{97.28}  & \textbf{89.44}&\textbf{97.99} &\textbf{62.51} &\textbf{87.67}\\
			\hline	
			Inception V3~\cite{hassannejad2016food}&  84.15& 96.11 &87.58&97.39&56.03&83.42\\			
			+Fine-tuned on Food2K& \textbf{87.61}& \textbf{97.25}& \textbf{89.89}&\textbf{98.20} &\textbf{62.48}&\textbf{87.71} \\
			\hline		
			DenseNet161~\cite{huang2017densely}& 86.94 &97.03 &86.93 & 97.17& 60.05&86.09\\		
			+Fine-tuned on Food2K& \textbf{88.22} & \textbf{97.57}&\textbf{89.54}&\textbf{97.84}&\textbf{62.46} &\textbf{87.75}\\
			\hline					
			SENet154~\cite{Jie2017Squeeze}& 88.62 &97.57  &88.71 &97.74 &63.83&88.61\\			
			+Fine-tuned on Food2K& \textbf{89.68}&\textbf{98.08} & \textbf{91.45}&\textbf{98.62} &\textbf{65.18}& \textbf{89.49}\\
			\hline	
			NTS-NET~(ResNet50)~\cite{Yang-L2Nav-ECCV2018}&89.40&97.80&89.24&97.91&64.61&89.45\\
			+Fine-tuned on Food2K& \textbf{90.78}& \textbf{98.17}&\textbf{90.33}&\textbf{98.37}&\textbf{65.34}&\textbf{89.53} \\
			\hline			
			Inception v4 @ 448px~\cite{Kornblith2018Do}&90.00&- &90.23 &97.86&65.49&88.35\\		
			+Fine-tuned on Food2K& \textbf{90.33} & \textbf{98.18}& \textbf{90.89}&\textbf{98.52}&\textbf{66.62}& \textbf{89.87}\\
			\hline	
			PAR-Net~(ResNet101)~\cite{Qiu-MDFR-BMVC2019}&89.30&- &89.60&-&63.98&88.76\\		
			+Fine-tuned on Food2K&  \textbf{89.93}& \textbf{97.72}& \textbf{90.92}&\textbf{98.74}&\textbf{64.72}&\textbf{89.23} \\
			\hline
			PMG(ResNet50)~\cite{du2020fine}&86.93&97.21&89.78&97.45&57.53&84.18\\
			+Fine-tuned on Food2K&\textbf{87.45}&\textbf{97.67}&\textbf{90.32}&\textbf{97.51}&\textbf{63.21}&\textbf{87.94}\\
			\hline
			DCL(ResNet50)~\cite{Chen_2019_CVPR}&88.90&97.82&89.13&97.82&64.10&88.77\\
			+Fine-tuned on Food2K&\textbf{89.57}&\textbf{97.85}&\textbf{90.80}&\textbf{98.41}&\textbf{64.37}&\textbf{88.90}\\
			\hline
		\end{tabular}
	\end{center}
	\label{Transfer-recognition}
\end{table*}

\subsubsection{Food Image Retrieval}
\begin{table*}[!t]
	\small
	\caption{Evaluating  visual representation  learned from ETH Food101 and Food2K for  retrieval on three datasets (\%). }
	\begin{center}
		\begin{tabular}{lcccccc}
			\hline
			\multirow{2}*{\textbf{Models}} &
			\multicolumn{2}{c}{ETH Food-101}&\multicolumn{2}{c}{Vireo Food-172}&\multicolumn{2}{c}{ISIA Food-500} \\
			\cline{2-7}& mAP & Recall@1& mAP & Recall@1 &mAP& Recall@1\\			
			\hline			
			Cross-entropy loss(ResNet101)&52.13&70.35&54.20&81.54&7.88 &36.58\\
			+Fine-tuned on ETH Food-101 &-&-&56.36&80.08&10.66&38.91\\
			+Fine-tuned on Food2K      &\textbf{53.33}&\textbf{78.10}&\textbf{68.23}&\textbf{85.87}&\textbf{11.20}&\textbf{40.04}\\
			\hline			
			Contrastive loss(ResNet101)~\cite{Hadsell-DR-CVPR06}&71.31&80.02&82.33&86.62&16.77&46.98\\
			+Fine-tuned on ETH Food-101 &-&-&76.62&83.53&16.42&46.02\\
			+Fine-tuned on Food2K                               &\textbf{77.91}&\textbf{83.30}&\textbf{83.61}&\textbf{87.07}&\textbf{21.52}&\textbf{49.64}\\
			\hline
			Triplet loss(ResNet101)~\cite{Schroff-FaceNet-CVPR05}&73.85&80.79&83.14&86.74&22.66&48.90\\	
			+Fine-tuned on ETH Food-101 &-&-&77.68&83.66&20.17&46.88\\
			+Fine-tuned on Food2K                                &\textbf{78.17}&\textbf{83.05}&\textbf{83.65}&\textbf{87.31}&\textbf{27.08}&\textbf{51.03}\\
			\hline				
		\end{tabular}
	\end{center}
	\label{transfer_retrieval}
\end{table*}
We further validate the generalization ability of Food2K on food image retrieval. Such task is to find food images containing relevant content with the given query in a database. We conduct the evaluation on ETH Food-101, Vireo Food-172 and ISIA Food-500 with the same training-test splitting in the mentioned papers. Each image in the test set is used in turn as the query, and the retrieval set is formed as all the remaining images of the test set. We initialize the networks using convolutional layers of ResNet101. The models are optimized using Adam.  The initial learning rate  $l_0 = 2 \times 10^{-5}$, an exponential decay ${l_0}exp(-0.1i)$ over epoch $i$, momentum 0.9, weight decay $5 \times 10^{-4}$, margin $\tau=0.85$ for  contrastive loss and triplet loss, and the batch size is 32. All training images are resized to a maximum size of $362 \times 362$, while keeping the original aspect ratio~\cite{Radenovic-FTCNN-TPAMI2019}. mAP and Recall@1 are  used as the evaluation metrics. The following methods are used: fine-tuning the network using Cross-entropy loss and metric-learning based methods using contrastive loss\cite{Hadsell-DR-CVPR06} and triplet loss \cite{Schroff-FaceNet-CVPR05}, respectively. ResNet101 is used as the backbone network for all methods. 

The results in Table \ref{transfer_retrieval} show that the methods using the backbone fine-tuned from Food2K all achieve the performance gain with various degrees on these benchmarks.  The improvement in Vireo Food-172 is the highest: the average improvement on these methods is 4.04\%, 5.28\% and 4.16\% in mAP for ETH Food-101, Vireo Food-172 and ISIA Food-500, which is consistent with the performance trend in food recognition. Further observation on Vireo Food-172 shows that when the performance from the base model is very high (probably close to saturation), the improvement from the model fine-tuned on Food2K is relatively small even the domain gap is small. For example, there is only 0.51\% improvement for the triplet loss in Vireo Food-172. In contrast, 4.32\% improvement is obtained in ETH Food-101. Note that there is no performance  improvement using additional ETH Food-101 for  metric-learning methods in our context. 
We speculate that metric-learning methods are more sensitive to the larger domain gap between  ETH Food-101 and these two target datasets~(especially Vireo Food-172), which indirectly indicates the higher diversity of categories and scale of images from Food2K enable better and stable generalization on food image retrieval. 
\subsubsection{Cross-Modal Recipe Retrieval}
We  evaluate the generalization of  Food2K on cross-modal recipe retrieval on Recipe1M, which is currently one popular task in the computer vision community~\cite{salvador2017learning}. We adopt the original data splits with 238,999 image-recipe pairs, 51,119 pairs and 51,303 pairs for training, validation and testing, respectively, and similar experiment setup~\cite{salvador2017learning}.  The evaluation metrics are the median retrieval rank (MedR) and the recall percentage at top K (Recall@K), i.e., the percentage of queries for which the matching answer is included in the top K results (K=1,5,10). 

Table \ref{transfer_cross} reports experimental results of different methods using backbones from ImageNet, ETH Food-101 and Food2K, and shows that  (1) all the methods obtains the improvement on the  model from ETH Food-101 and Food2K in MedR and R@K, and (2) The performance gain from Food2K is higher than ETH Food-101. These experimental results  prove that the backbone trained on Food2K is more helpful in visual food embedding learning, and thus improves cross-modal embedding learning. This is because of the diversity and scale of Food2K. Note that for JE~\cite{salvador2017learning} on 10k test size, because the original literature only uses the MedR for comparison, we also  adopt this metric  for consistent comparison. 

\begin{table*}[!t]
	\caption{Evaluating  visual representation  learned from ETH Food-101 and Food2K for  cross-modal recipe retrieval on the Recipe1M dataset (\%).}
	\begin{center}
		\begin{tabular}{cccccccccc}
			\hline
			\multirow{2}*{\textbf{Size}}&\multirow{2}*{\textbf{Methods}} &\multicolumn{4}{c}{Image-to-Recipe}&\multicolumn{4}{c}{Recipe-to-Image}\\
			\cline{3-10}&&MedR & Recall@1&Recall@5&Recall@10&MedR & Recall@1&Recall@5&Recall@10\\			
			\hline	
			\multirow{10}*{\textbf{1k}}&JE~\cite{salvador2017learning}& 5.2&24.0&51.0&65.0&5.1&25.0&52.0&65.0\\
			&+Fine-tuned on ETH Food-101&4.8&24.3&52.1&\textbf{66.2}&4.6&24.7&53.6&66.6\\
			&+Fine-tuned on Food2K&\textbf{4.3} &\textbf{25.8}&\textbf{53.3}&\textbf{66.2}&\textbf{4.1}&\textbf{26.4}&\textbf{54.5}&\textbf{66.9}\\	
			\cline{2-10}&VSE~\cite{weinberger2009distance}&3.9& 26.6&54.8 &68.2& 3.8&26.0 &55.7 &67.3 \\
			&+Fine-tuned on ETH Food-101&3.4&26.9&55.7&69.3&3.4&27.2&57.1&68.4\\
			&+Fine-tuned on Food2K&\textbf{2.8} &\textbf{29.1}&\textbf{58.5}&\textbf{70.6}&\textbf{2.9}&\textbf{28.6}&\textbf{58.6}&\textbf{70.5}\\	
			\cline{2-10}&PWC++~\cite{carvalho2018cross}&3.3& 25.8&54.5 &67.1 &3.5 &24.8 &55.0 &67.1 \\
			&+Fine-tuned on ETH Food-101&2.8&26.3&55.1&68.2&2.9&26.1&55.6&68.2\\
			&+Fine-tuned on Food2K&\textbf{2.6}&\textbf{27.4}&\textbf{56.3}&\textbf{69.3}&\textbf{2.7}&\textbf{27.1}&\textbf{56.7}&\textbf{68.8}\\	
			\cline{2-10}&AdaMine~\cite{carvalho2018cross}&2.5&39.8&69.0&77.4&2.1&40.2&68.1&78.7\\
			&+Fine-tuned on ETH Food-101&2.3&38.8&70.2&77.7&2.0&39.7&68.6&79.3\\
			&+Fine-tuned on Food2K&\textbf{1.9} &\textbf{41.2}&\textbf{71.3}&\textbf{78.5}&\textbf{1.8}&\textbf{40.5}&\textbf{69.6}&\textbf{79.9}\\	
			\cline{2-10}&ACME~\cite{wang2019learning}& 1.0&51.8&80.2&87.5&1.0&52.8&80.2&87.6\\	
			&+Fine-tuned on ETH Food-101&1.0&51.9&80.8&87.3&1.0&53.3&79.8&87.6\\
			&+Fine-tuned on Food2K&\textbf{1.0}&\textbf{52.3}&\textbf{81.3}&\textbf{88.9}&\textbf{1.0}&\textbf{53.7}&\textbf{80.9}&\textbf{88.3}\\
			\hline
			&JE~\cite{salvador2017learning}&41.9&-&-&-&39.2&-&-&-\\
			&+Fine-tuned on ETH Food-101&39.6&-&-&-&38.1&-&-&-\\
			&+Fine-tuned on Food2K&\textbf{37.5}&-&-&-&\textbf{35.2}&-&-&-\\	
			\cline{2-10}\multirow{10}*{\textbf{10k}}&VSE~\cite{weinberger2009distance}&28.3&7.4&21.5&29.6&28.1&7.6&22.4& 31.2\\
			&+Fine-tuned on ETH Food-101&27.1&8.6&22.3&30.7&27.3&8.5&24.1&32.7\\
			&+Fine-tuned on Food2K&\textbf{26.8}&\textbf{9.2}&\textbf{23.1}&\textbf{32.5}&\textbf{26.6}&\textbf{9.0}&\textbf{24.8}&\textbf{34.3}\\	
			\cline{2-10}&PWC++~\cite{carvalho2018cross}&27.5&7.4&20.3&29.9&27.6&7.7&21.3& 30.3\\
			&+Fine-tuned on ETH Food-101&26.4&7.9&21.8&30.8&26.8&8.3&22.4&31.4\\
			&+Fine-tuned on Food2K&\textbf{25.8}&\textbf{8.5}&\textbf{22.4}&\textbf{33.8}&\textbf{25.2}&\textbf{8.9}&\textbf{24.0}&\textbf{33.2}\\	
			\cline{2-10}&AdaMine~\cite{carvalho2018cross}&13.2&14.9&35.3&45.2&12.2&14.8&34.6&46.1\\
			&+Fine-tuned on ETH Food-101&12.5&15.2&36.1&46.9&11.7&15.4&35.7&46.7\\
			&+Fine-tuned on Food2K&\textbf{11.4}&\textbf{15.8}&\textbf{37.3}&\textbf{48.2}&\textbf{10.9}&\textbf{16.5}&\textbf{36.5}&\textbf{48.2}\\		
			\cline{2-10}&ACME~\cite{wang2019learning}&6.7&22.9&46.8&57.9&6.0&24.4&47.9&59.0\\
			&+Fine-tuned on ETH Food-101&6.5&23.2&47.1&58.3&5.4&23.9&48.0&59.8\\
			&+Fine-tuned on Food2K&\textbf{6.2}&\textbf{23.5}&\textbf{47.7}&\textbf{59.4}&\textbf{5.2}&\textbf{25.6}&\textbf{48.3}&\textbf{60.7}\\		
			\hline
		\end{tabular}
	\end{center}
	\label{transfer_cross}
\end{table*}

\begin{table*}[htbp]
	\small
	\caption{Evaluating  visual representation  learned from ETH Food-101 and Food2K for food detection on UNIMIB2016 and Oktoberfest(\%).}
	\begin{center}
		\begin{tabular}{lcccccc}
			\hline
			\multirow{2}*{\textbf{Models}} &\multicolumn{3}{c}{UNIMIB2016}&\multicolumn{3}{c}{Oktoberfest}\\
			\cline{2-7}&mAP & AP50 &AP75&mAP & AP50 &AP75\\			
			\hline
			SSD(VGG16)~\cite{liu2016ssd}& 64.1&78.4  &65.2 &52.4&79.8&54.8\\
			+Fine-tuned on ETH Food-101&64.3 &76.5 &73.2&54.8&80.0&62.1\\		
			+Fine-tuned on Food2K&\textbf{65.0} &\textbf{86.6} &\textbf{76.7}&\textbf{55.2}&\textbf{80.2}&\textbf{63.4} \\		
			\hline
			RetinaNet(ResNet101)~\cite{lin2017focal}& 66.8 &80.2&67.2&55.4&82.5&57.2 \\
			+Fine-tuned on ETH Food-101&67.3 &76.5 &70.9&55.7&78.2&58.9\\
			+Fine-tuned on Food2K&\textbf{69.1}&\textbf{80.9} &\textbf{76.1}&\textbf{55.8}&\textbf{84.7}&\textbf{59.0}\\	
			\hline	
			Faster-RCNN(ResNet101)~\cite{ren2015faster}& 55.2 &80.7&59.8&49.1&77.7&52.5\\
			+Fine-tuned on ETH Food-101&55.5&80.5 &60.3&51.8&83.0&52.9\\
			+Fine-tuned on Food2K&\textbf{55.7}&\textbf{82.8}&\textbf{60.4}&\textbf{57.6}&\textbf{83.6}&\textbf{64.0}\\	
			\hline
			PAN(ResNet101)~\cite{liu2018path}  & 58.3 & 88.6 & 66.3&50.4&77.8&56.5 \\
			+Fine-tuned on ETH Food-101&63.5 &85.2 &72.1&52.8&\textbf{81.8}&\textbf{58.3}\\
			+Fine-tuned on Food2K&\textbf{64.1}&\textbf{87.4}&\textbf{74.6}&\textbf{52.9}&{79.0}&\textbf{58.3}\\		
			\hline
			Cascade-RCNN(ResNet101)~\cite{cai2018cascade}& 73.1 &87.3&82.1&59.7&85.6&59.6\\	
			+Fine-tuned on ETH Food-101&73.2 &86.1 &82.7&61.2&85.9&70.1\\
			+Fine-tuned on Food2K&\textbf{73.7}&\textbf{87.5}&\textbf{83.2} &\textbf{61.7}&\textbf{87.8}&\textbf{70.2}\\	
			\hline
			Dynamic R-CNN(ResNet101)~\cite{zhang2020dynamic}& 68.8&90.6&77.9 &57.3&85.6&59.8\\
			+Fine-tuned on ETH Food-101&69.2 &88.3 &79.8&58.1&86.1&65.4\\
			+Fine-tuned on Food2K&\textbf{70.6}&\textbf{91.2}&\textbf{80.1} &\textbf{58.5}&\textbf{86.3}&\textbf{65.7}\\		
			\hline			
		\end{tabular}
	\end{center}
	\label{transfer_detection}
\end{table*}
\subsubsection{Food Detection}
\begin{figure}
	\centering
	\includegraphics[width=0.48\textwidth]{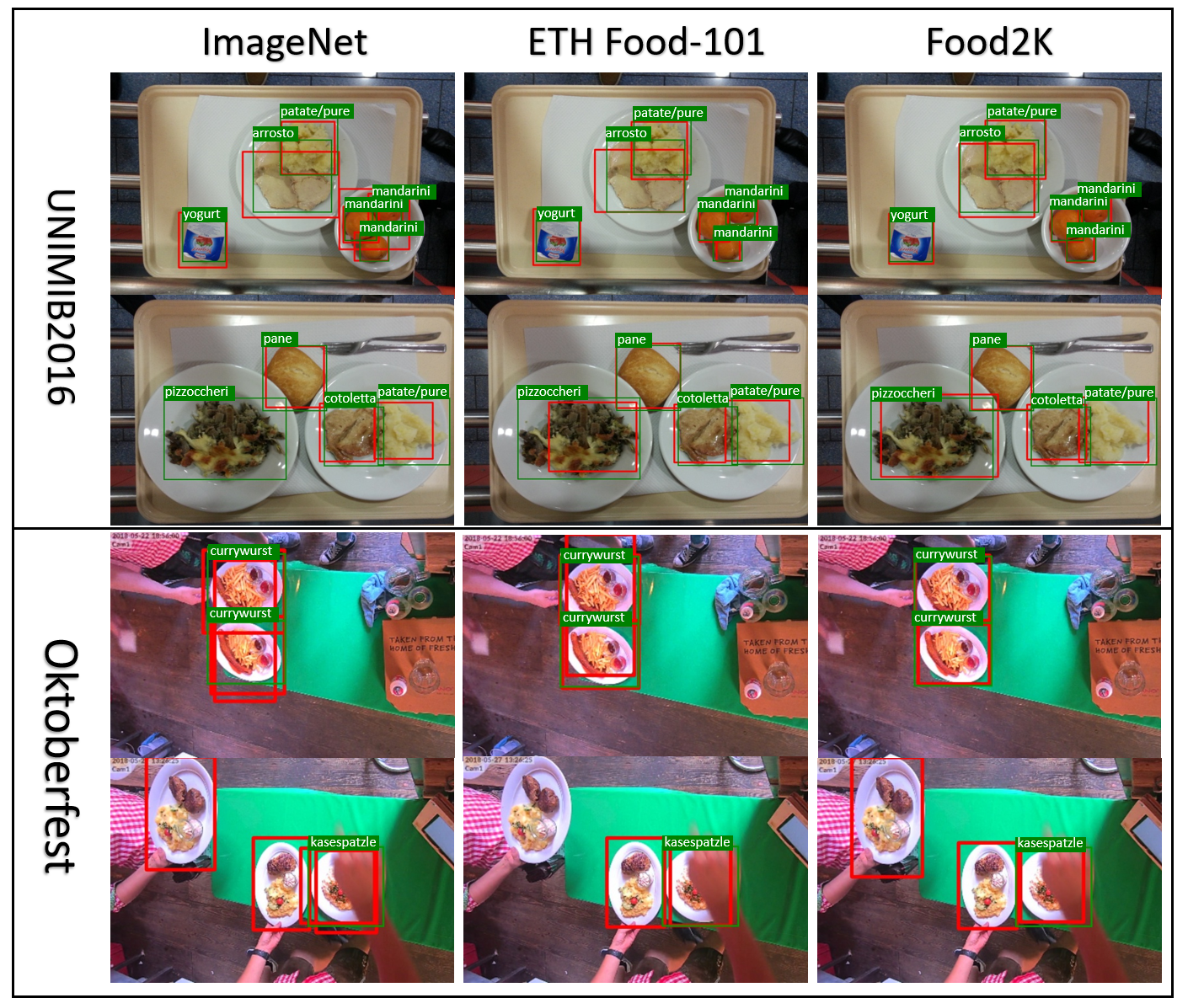}
	\caption{Comparison of food detection results via Dynamic R-CNN trained on ETH Food-101 and Food2K.}
	\label{detection_visi}
\end{figure}

We assess the generalization of models from Food2K on food detection, where the task is to detect food items from food trays. For comparison, we also conduct the evaluation on models trained from ETH Food-101 on this task. As the food detection model is supposed to have the ability of detecting negative samples (e.g., background), we add 2,500 non-food instance samples from~\cite{Singla-FnFC-MM2016} to  ETH Food-101 and Food2K  to fine-tune backbone models for detection. We conduct the evaluation on two available  datasets UNIMIB2016~\cite{Ciocca2016Food} and Oktoberfest~\cite{ziller2019oktoberfest} with original training-test splitting. Fig.~\ref{detection_visi} shows some examples. We train all the detectors using stochastic gradient descent with a batch-size of 8. The learning rate is  $10^{-3}$ and $5 \times 10^{-3}$ on UNIMIB2016 and Oktoberfest, respectively, and the input size of image is set to $512\times512$. The COCO style mAP, AP50 and AP75 are adopted as metrics, where the settings of IoU (Intersection over Union) threshhold for mAP, AP50 and AP75 are 0.50:.05:.95, 0.5 and 0.75 respectively.  Two single-stage (SSD~\cite{liu2016ssd}, RetinalNet~\cite{lin2017focal}) and four two-stage detection models (Faster-RCNN~\cite{ren2015faster},PAN~\cite{liu2018path},Cascade-RCNN~\cite{cai2018cascade}, Dynamic R-CNN~\cite{zhang2020dynamic}) with the same backbones mentioned in their papers are used for evaluation.

Table \ref{transfer_detection} reports the detection results of different methods using backbones from ImageNet, ETH Food-101 and Food2K, and shows that (1) all the methods obtain the improvement on the model from ETH Food-101 and Food2K in mAP and AP75, and (2) the performance gain from Food2K is higher than ETH Food-101, indicating the advantage of Food2K in both categories and image numbers. Particularly, the average mAP of all detectors from Food2K on UNIMIB2016 (66.4\%) is  higher (0.9\%) than it from ETH Food-101 (65.5\%). Similar trend can be found on Oktoberfest. It can be explained that
this is because of  the diversity  of  food categories  present in Food2K. With the supplement of more new food categories from Food2K, the trained backbones can help generalize better to food detection. In addition, considering UNIMIB2016 provides  more accurate annotations than Oktoberfest, and the environment background of UNIMIB2016 is more steady than Oktoberfest, it makes the backbone that delivers  better visual food features for detectors to get more obvious advantages in judging the categories of food instances after selecting positive proposals from the background, resulting in better performance gain on UNIMIB2016 than Oktoberfest. In addition, since  two-stage models depends more on the proposals based on the accurately extracted features, they obtains bigger increase in precision than two single-stage detectors. The average mAP growth of four two-stage models is 2.1\%  and the average mAP growth of two single-stage models is 1.6\% on two target datasets. It confirms that more discriminative features from better backbones can provide more obvious performance gain in two-stage detectors while the single-stage detectors relatively rely more on the structure of detection head.

We then visualize some detection results in Fig.~\ref{detection_visi}, and find that detectors with backbones from Food2K have better performance in providing high-quality proposals for all  methods. Meanwhile, representative examples chosen out of test samples in Oktoberfest prove that the reasons for relatively lower AP50 for some RCNN detectors (e.g., PAN~\cite{liu2018path}) from Food2K is that the models that learn abundant food features tend to recognize all the food instances in images even they are wrongly mislabeled and AP50 with IoU threshold 0.5 magnifies this "precise" flaw as it is only more tolerant of low-quality proposals and cannot reflect the models with high precision as mentioned in~\cite{cai2018cascade}. Therefore, several models have higher mAP and AP75 but lower AP50 can still be regarded as  more accurate detectors.

\subsubsection{Food Segmentation}
We  assess the generalization ability of  Food2K on food image segmentation  on the recently released  UEC-FoodPix Complete dataset~\cite{ege2019new}. It consists of 9,000 training images and 1,000 testing ones. The mean Intersection over Union (mIOU) and Pixel Accuracy (Pix Acc) are employed to evaluate the performance, where mIoU is a standard measurement for semantic segmentation that evaluates the overlap and the union in inference and ground truth, and Pix Acc is a  simpler measurement that is the accuracy for all pixels. Various models including FCN~\cite{long2015fully}, SegNet~\cite{badrinarayanan2017segnet}, PSPNet~\cite{zhao2017pyramid}, DUC\_HDC~\cite{wang2018understanding}, GCN~\cite{peng2017large} and DeepLabv3$+$~\cite{chen2018encoder} are adopted for evaluation. We employ the same learning rate schedule("poly" policy, the momentum 0.9 and same initial learning rate 0.01), crop size is $400 \times 400$, fine-tuning batch normalization parameters when output stride is 16, and random scale data augmentation during training. For backbones, we similarly add 2,500 non-food instance samples from~\cite{Singla-FnFC-MM2016} to  ETH Food-101 and Food2K  to make fine-tuned backbone models have the ability of segmenting negative samples.
\begin{table}[!t]
	\small
	\caption{Evaluating  visual representation  learned from ETH Food-101 and Food2K for food segmentation on the UEC-FoodPix Complete dataset (\%).}
	\begin{center}
		\begin{tabular}{lcc}
			\hline
			\multirow{2}*{\textbf{Models}} &\multicolumn{2}{c}{UECFoodPix Complete}\\
			\cline{2-3}&mIoU& PixAcc(PA)\\			
			\hline
			FCN8(ResNet50)~\cite{long2015fully}&55.8&70.2\\
			+Fine-tuned on ETH Food-101&56.5&\textbf{71.4}\\
			+Fine-tuned on Food2K&\textbf{56.6}&71.2\\		
			\hline
			SegNet(ResNet50)~\cite{badrinarayanan2017segnet}&57.6&71.7 \\
			+Fine-tuned on ETH Food-101&58.6&72.1\\
			+Fine-tuned on Food2K&\textbf{60.3}&\textbf{73.0}\\	
			\hline	
			PSPnet(ResNet50)~\cite{zhao2017pyramid}&70.9&81.7\\
			+Fine-tuned on ETH Food-101&72.2&82.0\\
			+Fine-tuned on Food2K&\textbf{74.5}&\textbf{84.1}\\	
			\hline
			DUC\_HDC(ResNet101)~\cite{wang2018understanding}  & 70.9&81.1  \\
			+Fine-tuned on ETH Food-101&{70.4}&{81.4}\\
			+Fine-tuned on Food2K&\textbf{72.7}&\textbf{82.5}\\		
			\hline
			Deeplabv3+(ResNet101)~\cite{chen2018encoder}& 71.2&82.4\\
			+Fine-tuned on ETH Food-101&71.6&82.5\\
			+Fine-tuned on Food2K&\textbf{73.4}&\textbf{83.3}\\		
			\hline			
		\end{tabular}
	\end{center}
	\label{transfer_segmentation}
\end{table}
\begin{figure}
	\centering
	\includegraphics[width=0.48\textwidth]{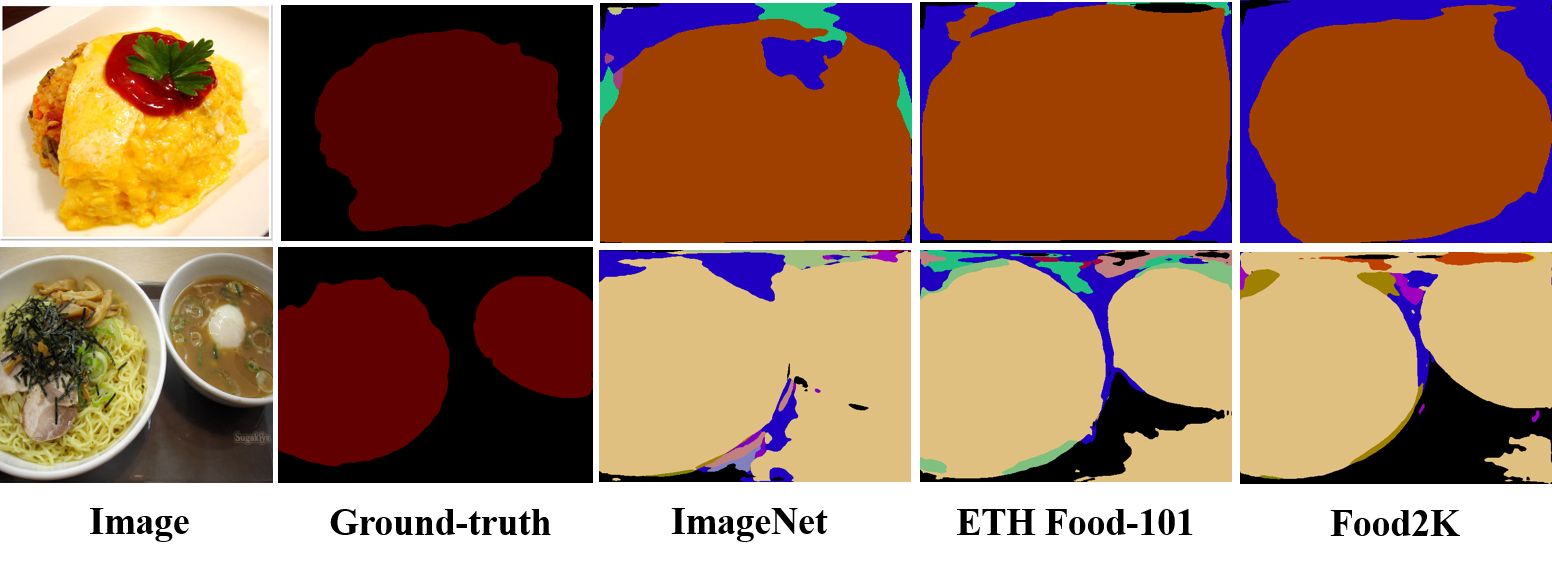}
	\caption{Comparison of food segmentation results via DeeplabV3 trained on ETH Food-101 and Food2K.}
	\label{segmentVis}
\end{figure}

Table \ref{transfer_segmentation} reports segmentation results of different methods. We can see that our algorithm can provide higher performance gain by training the backbone part using Food2K for most of methods. For example, we have more than 2 points improvement over ImageNet pre-trained  models for DeepLabv3+ model and also  improvement over ETH Food-101. We also provide visual comparison results with DeeplabV3 on UEC-FoodPix Complete in Fig.\ref{segmentVis}. We can see that our model can segment food regions more accurately than others.





\subsection{Discussion}
Our final set of experiments demonstrate the generality of the learned features from Food2K for various vision and multimodal tasks, indicating its usefulness and value. We believe this is because of the higher diversity and larger scale of Food2K. Below we discuss  some  potential research problems and methods based on Food2K.

(1) \textbf{Large-scale robust food recognition} Based on our experimental results, although existing fine-grained recognition methods, e.g.,  PMG~\cite{du2020fine} obtain the state-of-the art performance in existing fine-grained datasets, they fail to obtain the desired  performance on Food2K. In addition, there are also some recently proposed food recognition methods, such as PAR-Net~\cite{Qiu-MDFR-BMVC2019}, which have achieved better performance in small or median-level recognition datasets. However, they also fail to obtain better performance in large-scale food recognition on Food2K. We speculate that there are more complex visual patterns about food generated from different ingredients, accessories and arrangements with the increase of both the diversity and scale of food data, and these methods are not suitable or robust  for this case. As one initial attempt, we combined  progressive training and self-attention to learn more stable and discriminative global and local features, resulting in good performance.  More methods are worth further exploration. For example, recently, transformers have made tremendous impact in  image recognition~\cite{dosovitskiy2020image}, where the performance is higher  than CNNs on large-scale datasets. Food2K can provide sufficient training data to develop transformer-based food recognition methods to improve its performance. 

(2)  \textbf{Human vision evaluation on food recognition} Conducting human vision research on Food2K is also an interesting topic to study. Compared with  human vision research on generic object recognition, it is probably more difficult  to conduct such evaluation on food recognition. For example, food has strong regional and cultural characteristics, and human subjects from different regions thus  have stronger bias for food recognition. Recent works~\cite{Furtado-HVS-CVIU2020} give an initial empirical comparison between human visual system and CNNs in the food recognition task. In order to avoid information overburden, the number of dishes to learn was restricted to 16 different types of food for human subjects. More interesting problems can be further explored. For example,  What is the upper bound for human performance on food recognition? What's their own  advantages and disadvantages for human vision 
system and CNNs in recognizing food types and number of categories? Moreover, knowledge from other fields, e.g., food science is probably needed for further explanation  on  experimental results.

(3) \textbf{Cross-X transfer learning for food recognition} We have verified the generalization of Food2K in various vision and multimodal tasks. We can study the transfer learning from more aspects in the future. For example, food has its own geographical and culture attributes. We can conduct cross-cuisine transfer learning. That means we use trained models from eastern cuisines for performance analysis on western cuisines, and vice versa. After more fine-grained scenario annotation, such as region-level or even restaurant-level annotation, we can  conduct cross-scenario transfer learning for food recognition. In addition, we can also study cross super-class transfer learning for food recognition. For example,  we can  use trained models from the seafood super-class for performance analysis on the meat super-class. These interesting problems are worth deep exploration.

(4) \textbf{Large-Scale Few-Shot Food Recognition~(LS-FSFR)}  Recently, there are some works on few-shot food recognition on small/medium-scale food categories ~\cite{Shuqiang-FSFR-TOMM2020,Zhao-FL-WACV2021}. In contrast, LS-FSFR is a more realistic task that aims to identify hundreds of novel food categories without forgetting those categories, where each novel category has only a few samples~\cite{Aoxue-LSFSL-CVPR2019}. Food2K provides such large-scale food dataset test benchmark  to support this task. In addition, the constructed food ontology can also  help the method design of LS-FSFR as the prior knowledge.

(5) \textbf{More applications on Food2K} We have verified better generalization ability of Food2K in various tasks, including food image recognition, food image retrieval, cross-modal recipe retrieval, food detection and segmentation in the paper.  Furthermore, Food2K can also support more novel applications. Food image generation is one novel and interesting applications, and it can synthesize new food images which are similar to those in real-life scenarios by Generative Adversarial Networks (GANs)~\cite{goodfellow2014generative}. For example, Zhu \textit{et al.}~\cite{Zhu-CookGAN-CVPR2020} can generate  highly realistic and semantically consistent images from given ingredients and instructions. Another work~\cite{Papadopoulos-MakePizza-CVPR2019} aims to teach a machine how to make a pizza by building a generative model that mirrors the step-by-step procedure. Different GANs such as Lightweight GAN~\cite{liu2020towards} can also be used to generate synthetic food images based on Food2K. Please refer to the supplementary materials for more details about the evaluation for food image generation on Food2K.

(6) \textbf{Extension of Food2K for more tasks}  Researchers are encouraged to apply trained models on Food2K to more food-relevant tasks. Moreover, we hope Food2K will evolve over time. Considering that some works~\cite{Chen-DIRCRR-MM2016} have showed that ingredients can improve the recognition performance, we plan to extend Food2K by providing richer attribute annotation to support  food recognition with different semantic levels.  We can also  conduct region-level and pixel-level annotation on Food2K to enable broader range of its application. In addition, we can also conduct some novel tasks, such as aesthetic assessment of food images via annotating aesthetic attribute labels on Food2K~\cite{Kekai-Gourmet-SIGGRAPH2018}. 
\section{Conclusions}
\label{Conclusion}
In this paper, we present Food2K with larger data volume, larger category coverage and higher diversity compared with existing ones, which can  be served as a new benchmark for scalable  food recognition. It can benefit various vision and multimodal tasks, including food recognition, retrieval, detection, segmentation, and cross-modal recipe retrieval for  its better generalization ability. To date, Food2K is the largest food recognition dataset with its diversity and scale. We believe it will enable development of large-scale food recognition methods, and also help the researchers to utilize  Food2K  for the future research on  more food-relevant tasks, such as large-scale few-shot food recognition and  transfer learning on food recognition from various aspects, such as cross-scenario, cross-cuisine and cross-super-class transfer learning.

%
%

\ifCLASSOPTIONcaptionsoff
  \newpage
\fi
%
%
%
\bibliographystyle{IEEEtran}
\normalem
\bibliography{Food2k.bib}

\begin{thebibliography}{100}
\providecommand{\url}[1]{#1}
\csname url@samestyle\endcsname
\providecommand{\newblock}{\relax}
\providecommand{\bibinfo}[2]{#2}
\providecommand{\BIBentrySTDinterwordspacing}{\spaceskip=0pt\relax}
\providecommand{\BIBentryALTinterwordstretchfactor}{4}
\providecommand{\BIBentryALTinterwordspacing}{\spaceskip=\fontdimen2\font plus
\BIBentryALTinterwordstretchfactor\fontdimen3\font minus
  \fontdimen4\font\relax}
\providecommand{\BIBforeignlanguage}[2]{{%
\expandafter\ifx\csname l@#1\endcsname\relax
\typeout{** WARNING: IEEEtran.bst: No hyphenation pattern has been}%
\typeout{** loaded for the language `#1'. Using the pattern for}%
\typeout{** the default language instead.}%
\else
\language=\csname l@#1\endcsname
\fi
#2}}
\providecommand{\BIBdecl}{\relax}
\BIBdecl

\bibitem{Min2019A}
W.~Min, S.~Jiang, L.~Liu, Y.~Rui, and R.~Jain, ``A survey on food computing,''
  \emph{ACM CSUR}, vol.~52, no.~5, pp. 1--36, 2019.

\bibitem{Boswell-FC-PNAS2018}
R.~G. Boswell, W.~Sun, S.~Suzuki, and H.~Kober, ``Training in cognitive
  strategies reduces eating and improves food choice,'' \emph{PNAS}, vol. 115,
  no.~48, pp. E11\,238--E11\,247, 2018.

\bibitem{David-DH-Nature2014}
T.~David and C.~Michael, ``Global diets link environmental sustainability and
  human health,'' \emph{Nature}, vol. 515, no. 7528, pp. 518--22, 2014.

\bibitem{Rozin-SFRHO-SB1976}
P.~Rozin, ``The selection of foods by rats, humans, and other animals,'' ser.
  Advances in the Study of Behavior, 1976, vol.~6, pp. 21--76.

\bibitem{Meyers-Im2Calories-ICCV2015}
A.~Meyers, N.~Johnston, V.~Rathod, A.~Korattikara, A.~Gorban, N.~Silberman,
  S.~Guadarrama, G.~Papandreou, J.~Huang, and K.~P. Murphy, ``{Im2Calories}:
  towards an automated mobile vision food diary,'' in \emph{ICCV}, 2015, pp.
  1233--1241.

\bibitem{Quin-Nutrition5k-CVPR2021}
Q.~Thames, A.~Karpur, W.~Norris, F.~Xia, L.~Panait, T.~Weyand, and J.~Sim,
  ``Nutrition5k: Towards automatic nutritional understanding of generic food,''
  in \emph{Proceedings of the IEEE/CVF Conference on Computer Vision and
  Pattern Recognition (CVPR)}, June 2021, pp. 8903--8911.

\bibitem{Lu-AIS-TMM2020}
Y.~{Lu}, T.~{Stathopoulou}, M.~F. {Vasiloglou}, S.~{Christodoulidis},
  Z.~{Stanga}, and S.~{Mougiakakou}, ``An artificial intelligence-based system
  to assess nutrient intake for hospitalised patients,'' \emph{IEEE TMM}, pp.
  1--1, 2020.

\bibitem{Bossard-Food101-ECCV2014}
L.~Bossard, M.~Guillaumin, and L.~Van~Gool, ``Food-101--mining discriminative
  components with random forests,'' in \emph{ECCV}, 2014, pp. 446--461.

\bibitem{Qiu-MDFR-BMVC2019}
J.~Qiu, F.~P.-W.~Lo, Y.~Sun, S.~Wang, and B.~Lo, ``Mining discriminative food
  regions for accurate food recognition,'' in \emph{BMVC}, 2019.

\bibitem{Martinel-WSR-WACV2018}
N.~Martinel, G.~L. Foresti, and C.~Micheloni, ``Wide-slice residual networks
  for food recognition,'' in \emph{WACV}, 2018, pp. 567--576.

\bibitem{Parneet-FoodX251-CVPRW2019}
P.~Kaur, K.~Sikka, W.~Wang, S.~J. Belongie, and A.~Divakaran, ``Foodx-251: {A}
  dataset for fine-grained food classification,'' in \emph{CVPRW}, 2019.

\bibitem{Min-MSMVFA-TIP2019}
S.~Jiang, W.~Min, L.~Liu, and Z.~Luo, ``Multi-scale multi-view deep feature
  aggregation for food recognition,'' \emph{IEEE TIP}, vol.~29, no.~1, pp.
  265--276, 2019.

\bibitem{Lixi2019Mixed}
L.~Deng, J.~Chen, Q.~Sun, X.~He, S.~Tang, Z.~Ming, Y.~Zhang, and T.~S. Chua,
  ``Mixed-dish recognition with contextual relation networks,'' in \emph{ACM
  MM}, 2019, pp. 112--120.

\bibitem{Zhao-FL-WACV2021}
H.~Zhao, K.-H. Yap, and A.~Chichung~Kot, ``Fusion learning using semantics and
  graph convolutional network for visual food recognition,'' in \emph{WACV},
  2021, pp. 1711--1720.

\bibitem{Chen-DIRCRR-MM2016}
J.~Chen and C.-W. Ngo, ``Deep-based ingredient recognition for cooking recipe
  retrieval,'' in \emph{ACM MM}, 2016, pp. 32--41.

\bibitem{Min-ISIA-500-MM2020}
W.~Min, L.~Liu, Z.~Wang, Z.~Luo, X.~Wei, and X.~Wei, ``{ISIA Food-500}: A
  dataset for large-scale food recognition via stacked global-local attention
  network,'' in \emph{ACM MM}, 2020, pp. 393--401.

\bibitem{Salvador-LCME-arXiv2018}
J.~Mar{\'{\i}}n, A.~Biswas, F.~Ofli, N.~Hynes, A.~Salvador, Y.~Aytar, I.~Weber,
  and A.~Torralba, ``{Recipe1M+}: {A} dataset for learning cross-modal
  embeddings for cooking recipes and food images,'' \emph{{IEEE} TPAMI},
  vol.~43, no.~1, pp. 187--203, 2021.

\bibitem{WangLHM20-SAGN-ECCV2020}
H.~Wang, G.~Lin, S.~C.~H. Hoi, and C.~Miao, ``Structure-aware generation
  network for recipe generation from images,'' in \emph{ECCV}, vol. 12372,
  2020, pp. 359--374.

\bibitem{Chen2009PFID}
M.~Chen, K.~Dhingra, W.~Wu, L.~Yang, R.~Sukthankar, and J.~Yang, ``{PFID}:
  Pittsburgh fast-food image dataset,'' in \emph{ICIP}, 2009, pp. 289--292.

\bibitem{Joutou2010A}
T.~Joutou and K.~Yanai, ``A food image recognition system with multiple kernel
  learning,'' in \emph{ICIP}, 2009, pp. 285--288.

\bibitem{Hoashi2010Image}
H.~Hoashi, T.~Joutou, and K.~Yanai, ``Image recognition of 85 food categories
  by feature fusion,'' in \emph{ISM}, 2010, pp. 296--301.

\bibitem{Matsuda2012Multiple}
Y.~Matsuda and K.~Yanai, ``Multiple-food recognition considering co-occurrence
  employing manifold ranking,'' in \emph{ICPR}, 2012, pp. 2017--2020.

\bibitem{kawano2014automatic}
Y.~Kawano and K.~Yanai, ``Automatic expansion of a food image dataset
  leveraging existing categories with domain adaptation,'' in \emph{ECCV},
  2014, pp. 3--17.

\bibitem{Anthimopoulos2014A}
M.~M. Anthimopoulos, L.~Gianola, L.~Scarnato, P.~Diem, and S.~G. Mougiakakou,
  ``A food recognition system for diabetic patients based on an optimized
  bag-of-features model,'' \emph{IEEE JBHI}, vol.~18, no.~4, pp. 1261--1271,
  2014.

\bibitem{Wang-RRLMFD-ICME2015}
X.~Wang, D.~Kumar, N.~Thome, M.~Cord, and F.~Precioso, ``Recipe recognition
  with large multimodal food dataset,'' in \emph{ICME}, 2015, pp. 1--6.

\bibitem{XuRuihan-GMDR-TMM2015}
R.~Xu, L.~Herranz, S.~Jiang, S.~Wang, X.~Song, and R.~Jain, ``Geolocalized
  modeling for dish recognition,'' \emph{IEEE TMM}, vol.~17, no.~8, pp.
  1187--1199, 2015.

\bibitem{Farinella2015A}
G.~M. Farinella, D.~Allegra, and F.~Stanco, ``A benchmark dataset to study the
  representation of food images,'' in \emph{ECCV}, 2014, pp. 584--599.

\bibitem{Zhou_FGIC_CVPR2016}
F.~Zhou and Y.~Lin, ``Fine-grained image classification by exploring
  bipartite-graph labels,'' in \emph{CVPR}, 2016, pp. 1124--1133.

\bibitem{Merler2016}
M.~Merler, H.~Wu, R.~Uceda-Sosa, Q.-B. Nguyen, and J.~R. Smith, ``Snap, eat,
  repeat: A food recognition engine for dietary logging,'' in \emph{MADiMa},
  2016, pp. 31--40.

\bibitem{Singla-FnFC-MM2016}
A.~Singla, L.~Yuan, and T.~Ebrahimi, ``Food/non-food image classification and
  food categorization using pre-trained googlenet model,'' in \emph{MADiMa},
  2016, pp. 3--11.

\bibitem{Farinella2016Retrieval}
G.~M. Farinella, D.~Allegra, M.~Moltisanti, F.~Stanco, and S.~Battiato,
  ``Retrieval and classification of food images,'' \emph{CBM}, vol.~77, pp.
  23--39, 2016.

\bibitem{ciocca2017learning}
G.~Ciocca, P.~Napoletano, and R.~Schettini, ``Learning {CNN}-based features for
  retrieval of food images,'' in \emph{ICIAP}, 2017, pp. 426--434.

\bibitem{Chen2017ChineseFoodNet}
X.~Chen, H.~Zhou, and L.~Diao, ``{ChineseFoodNet}: A large-scale image dataset
  for {Chinese} food recognition,'' \emph{CoRR}, vol. abs/1705.02743, 2017.

\bibitem{Hou-VegFru-ICCV2017}
S.~Hou, Y.~Feng, and Z.~Wang, ``{VegFru}: A domain-specific dataset for
  fine-grained visual categorization,'' in \emph{ICCV}, 2017, pp. 541--549.

\bibitem{Min-IGCMAN-ACMMM2019}
W.~Min, L.~Liu, Z.~Luo, and S.~Jiang, ``Ingredient-guided cascaded
  multi-attention network for food recognition,'' in \emph{ACM MM}, 2019, pp.
  99--107.

\bibitem{Doyen-FoodAI-KDD2019}
D.~Sahoo, W.~Hao, S.~Ke, X.~Wu, H.~Le, P.~Achananuparp, E.~Lim, and S.~C.~H.
  Hoi, ``{FoodAI}: Food image recognition via deep learning for smart food
  logging,'' in \emph{ACM KDD}, 2019, pp. 2260--2268.

\bibitem{Deng-ImageNet-CVPR2009}
J.~Deng, W.~Dong, R.~Socher, L.~Li, K.~Li, and F.~Li, ``{ImageNet}: A
  large-scale hierarchical image database,'' in \emph{CVPR}, 2009, pp.
  248--255.

\bibitem{Yanai-FIR-ICMEW2015}
K.~{Yanai} and Y.~{Kawano}, ``Food image recognition using deep convolutional
  network with pre-training and fine-tuning,'' in \emph{ICMEW}, 2015, pp. 1--6.

\bibitem{Min-YAWYE-TMM2018}
W.~Min, B.-K. Bao, S.~Mei, Y.~Zhu, Y.~Rui, and S.~Jiang, ``You are what you
  eat: Exploring rich recipe information for cross-region food analysis,''
  \emph{IEEE TMM}, vol.~20, no.~4, pp. 950--964, 2018.

\bibitem{Aguilar2018Grab}
E.~Aguilar, B.~Remeseiro, M.~Bola{\~n}os, and P.~Radeva, ``{Grab, Pay and Eat}:
  Semantic food detection for smart restaurants,'' \emph{IEEE TMM}, vol.~20,
  no.~12, pp. 3266--3275, 2018.

\bibitem{Yang-FR-CVPR2010}
S.~Yang, M.~Chen, D.~Pomerleau, and R.~Sukthankar, ``Food recognition using
  statistics of pairwise local features,'' in \emph{CVPR}, 2010, pp.
  2249--2256.

\bibitem{Martinel2016A}
N.~Martinel, C.~Piciarelli, and C.~Micheloni, ``A supervised extreme learning
  committee for food recognition,'' \emph{CVIU}, vol. 148, pp. 67--86, 2016.

\bibitem{Shuqiang-FSFR-TOMM2020}
S.~Jiang, W.~Min, Y.~Lyu, and L.~Liu, ``Few-shot food recognition via
  multi-view representation learning,'' \emph{{ACM} {TOMM}}, vol.~16, no.~3,
  pp. 87:1--87:20, 2020.

\bibitem{Beijbom-MeMa-WACV2015}
O.~Beijbom, N.~Joshi, D.~Morris, S.~Saponas, and S.~Khullar, ``Menu-match:
  restaurant-specific food logging from images,'' in \emph{WACV}, 2015, pp.
  844--851.

\bibitem{Horiguchi-PCFIR-TMM2018}
S.~Horiguchi, S.~Amano, M.~Ogawa, and K.~Aizawa, ``Personalized classifier for
  food image recognition,'' \emph{{IEEE} TMM}, vol.~20, no.~10, pp. 2836--2848,
  2018.

\bibitem{ege2019new}
K.~Okamoto and K.~Yanai, ``{UEC-FoodPix Complete}: {A} large-scale food image
  segmentation dataset,'' in \emph{{ICPRW}}, vol. 12665, 2021, pp. 647--659.

\bibitem{salvador2017learning}
A.~Salvador, N.~Hynes, Y.~Aytar, J.~Marin, F.~Ofli, I.~Weber, and A.~Torralba,
  ``Learning cross-modal embeddings for cooking recipes and food images,'' in
  \emph{CVPR}, 2017, pp. 3020--3028.

\bibitem{wang2019learning}
H.~Wang, D.~Sahoo, C.~Liu, E.-p. Lim, and S.~C. Hoi, ``Learning cross-modal
  embeddings with adversarial networks for cooking recipes and food images,''
  in \emph{CVPR}, 2019, pp. 11\,572--11\,581.

\bibitem{papadopoulos2022learning}
D.~P. Papadopoulos, E.~Mora, N.~Chepurko, K.~W. Huang, F.~Ofli, and
  A.~Torralba, ``Learning program representations for food images and cooking
  recipes,'' in \emph{Proceedings of the IEEE/CVF Conference on Computer Vision
  and Pattern Recognition}, 2022, pp. 16\,559--16\,569.

\bibitem{fu2020mcen}
H.~Fu, R.~Wu, C.~Liu, and J.~Sun, ``{MCEN}: Bridging cross-modal gap between
  cooking recipes and dish images with latent variable model,'' in \emph{CVPR},
  2020, pp. 14\,570--14\,580.

\bibitem{Salvador-HT-CVPR2021}
A.~Salvador, E.~Gundogdu, L.~Bazzani, and M.~Donoser, ``Revamping cross-modal
  recipe retrieval with hierarchical transformers and self-supervised
  learning,'' in \emph{CVPR}, 2021, pp. 15\,475--15\,484.

\bibitem{Papadopoulos-MakePizza-CVPR2019}
D.~P. Papadopoulos, Y.~Tamaazousti, F.~Ofli, I.~Weber, and A.~Torralba, ``How
  to make a pizza: Learning a compositional layer-based {GAN} model,'' in
  \emph{The IEEE Conference on Computer Vision and Pattern Recognition}, 2019.

\bibitem{Fangda-CookGAN-WACV2020}
F.~Han, R.~Guerrero, and V.~Pavlovic, ``{CookGAN}: Meal image synthesis from
  ingredients,'' in \emph{WACV}, 2020, pp. 1439--1447.

\bibitem{Zhu-CookGAN-CVPR2020}
B.~Zhu and C.~Ngo, ``{CookGAN}: Causality based text-to-image synthesis,'' in
  \emph{{CVPR}}, 2020, pp. 5518--5526.

\bibitem{salvador2019inverse}
A.~Salvador, M.~Drozdzal, X.~Gir{\'o}-i Nieto, and A.~Romero, ``Inverse
  cooking: Recipe generation from food images,'' in \emph{Proceedings of the
  IEEE/CVF Conference on Computer Vision and Pattern Recognition}, 2019, pp.
  10\,453--10\,462.

\bibitem{Salvador-IC-CVPR2019}
A.~Salvador, M.~Drozdzal, X.~{Gir{\'{o}} i Nieto}, and A.~Romero, ``Inverse
  cooking: Recipe generation from food images,'' in \emph{{CVPR}}, 2019, pp.
  10\,445--10\,454.

\bibitem{nestle2013food}
M.~Nestle, \emph{Food politics: How the food industry influences nutrition and
  health}, 2013, vol.~3.

\bibitem{food_safety}
``National food safety standard for uses of food additives (gb 31632-2014),''
  \emph{China Food Additives}, no.~8X, p.~28, 2015.

\bibitem{zhang2019learning}
L.~Zhang, S.~Huang, W.~Liu, and D.~Tao, ``Learning a mixture of
  granularity-specific experts for fine-grained categorization,'' in
  \emph{ICCV}, 2019, pp. 8331--8340.

\bibitem{wang2018non}
X.~Wang, R.~Girshick, A.~Gupta, and K.~He, ``Non-local neural networks,'' in
  \emph{CVPR}, 2018, pp. 7794--7803.

\bibitem{huang2017densely}
G.~Huang, Z.~Liu, K.~Q. Weinberger, and L.~van~der Maaten, ``Densely connected
  convolutional networks,'' in \emph{CVPR}, 2017, pp. 4700--4708.

\bibitem{Jie2017Squeeze}
J.~Hu, L.~Shen, and G.~Sun, ``Squeeze-and-excitation networks,'' in
  \emph{CVPR}, 2018, pp. 7132--7141.

\bibitem{2020Multi}
S.~Min, H.~Yao, H.~Xie, Z.-J. Zha, and Y.~Zhang, ``Multi-objective matrix
  normalization for fine-grained visual recognition,'' \emph{IEEE TIP},
  vol.~29, pp. 4996--5009, 2020.

\bibitem{du2020fine}
R.~Du, D.~Chang, A.~K. Bhunia, J.~Xie, Z.~Ma, Y.~Song, and J.~Guo,
  ``Fine-grained visual classification via progressive multi-granularity
  training of jigsaw patches,'' in \emph{{ECCV}}, 2020, pp. 153--168.

\bibitem{Szegedy-GDC-CVPR2015}
C.~Szegedy, W.~Liu, Y.~Jia, P.~Sermanet, S.~Reed, D.~Anguelov, D.~Erhan,
  V.~Vanhoucke, and A.~Rabinovich, ``Going deeper with convolutions,'' in
  \emph{CVPR}, 2015, pp. 1--9.

\bibitem{Szegedy-Inception-v4-AAAI2017}
C.~Szegedy, S.~Ioffe, V.~Vanhoucke, and A.~A. Alemi, ``Inception-v4,
  inception-resnet and the impact of residual connections on learning,'' in
  \emph{{AAAI}}, 2017, pp. 4278--4284.

\bibitem{He-DRL-CVPR2016}
K.~He, X.~Zhang, S.~Ren, and J.~Sun, ``Deep residual learning for image
  recognition,'' in \emph{CVPR}, 2016, pp. 770--778.

\bibitem{Zagoruyko-WRN-BMVC2016}
Z.~Sergey and K.~Nikos, ``Wide residual networks,'' in \emph{BMVC}, 2016, pp.
  87.1--87.12.

\bibitem{Yang-L2Nav-ECCV2018}
Z.~Yang, T.~Luo, D.~Wang, Z.~Hu, J.~Gao, and L.~Wang, ``Learning to navigate
  for fine-grained classification,'' in \emph{ECCV}, 2018, pp. 438--454.

\bibitem{Yu_2018_ECCV}
C.~Yu, X.~Zhao, Q.~Zheng, P.~Zhang, and X.~You, ``Hierarchical bilinear pooling
  for fine-grained visual recognition,'' in \emph{ECCV}, 2018, pp. 595--610.

\bibitem{Chen_2019_CVPR}
Y.~Chen, Y.~Bai, W.~Zhang, and T.~Mei, ``Destruction and construction learning
  for fine-grained image recognition,'' in \emph{CVPR}, 2019, pp. 5157--5166.

\bibitem{Hu_2019_CVPR}
T.~Hu, H.~Qi, Q.~Huang, and Y.~Lu, ``See better before looking closer: Weakly
  supervised data augmentation network for fine-grained visual
  classification,'' \emph{arXiv preprint arXiv:1901.09891}, 2019.

\bibitem{Kornblith2018Do}
S.~Kornblith, J.~Shlens, and Q.~Le, ``Do better {ImageNet} models transfer
  better?'' in \emph{CVPR}, 2019, pp. 2661--2671.

\bibitem{Yu-DLA-CVPR2018}
F.~Yu, D.~Wang, E.~Shelhamer, and T.~Darrell, ``Deep layer aggregation,'' in
  \emph{CVPR}, 2018, pp. 2403--2412.

\bibitem{mcallister2018combining}
P.~McAllister, H.~Zheng, R.~Bond, and A.~Moorhead, ``Combining deep residual
  neural network features with supervised machine learning algorithms to
  classify diverse food image datasets,'' \emph{Computers in Biology and
  Medicine}, vol.~95, pp. 217--233, 2018.

\bibitem{Yanai-FIRDCNN-ICME2015}
K.~Yanai and Y.~Kawano, ``Food image recognition using deep convolutional
  network with pre-training and fine-tuning,'' in \emph{ICMEW}, 2015, pp. 1--6.

\bibitem{Wu-LMBM-MM2016}
H.~Wu, M.~Merler, R.~Uceda-Sosa, and J.~R. Smith, ``Learning to make better
  mistakes: Semantics-aware visual food recognition,'' in \emph{ACM MM}, 2016,
  pp. 172--176.

\bibitem{Pandey2017FoodNet}
P.~Pandey, A.~Deepthi, B.~Mandal, and N.~B. Puhan, ``{FoodNet}: Recognizing
  foods using ensemble of deep networks,'' \emph{IEEE TSPL}, vol.~24, no.~12,
  pp. 1758--1762, 2017.

\bibitem{Ao2015Adapting}
S.~Ao and C.~X. Ling, ``Adapting new categories for food recognition with deep
  representation,'' in \emph{ICDMW}, 2015, pp. 1196--1203.

\bibitem{Liu2016DeepFood}
C.~Liu, Y.~Cao, Y.~Luo, G.~Chen, V.~Vokkarane, and Y.~Ma, ``Deepfood: Deep
  learning-based food image recognition for computer-aided dietary
  assessment,'' in \emph{Inclusive Smart Cities and Digital Health}, 2016, pp.
  37--48.

\bibitem{Bola2017Simultaneous}
M.~Bolanos and P.~Radeva, ``Simultaneous food localization and recognition,''
  \emph{ICPR}, pp. 3140--3145, 2017.

\bibitem{Rodriguez-PATA-TMM2019}
P.~R. López, D.~V. Dorta, G.~C. Preixens, J.~M. Gonfaus, and J.~G. Sabaté,
  ``Pay attention to the activations: a modular attention mechanism for
  fine-grained image recognition,'' \emph{IEEE TMM}, vol.~22, no.~2, pp.
  502--514, 2020.

\bibitem{Aguilar2017Food}
E.~Aguilar, M.~Bola{\~{n}}os, and P.~Radeva, ``Food recognition using fusion of
  classifiers based on cnns,'' in \emph{ICIAP}, 2017, pp. 213--224.

\bibitem{Hassanne-FIRDCN-MM2016}
H.~Hassannejad, G.~Matrella, P.~Ciampolini, I.~D. Munari, M.~Mordonini, and
  S.~Cagnoni, ``Food image recognition using very deep convolutional
  networks,'' in \emph{MADiMa}, 2016, pp. 41--49.

\bibitem{martinel2018wide}
N.~Martinel, G.~L. Foresti, and C.~Micheloni, ``Wide-slice residual networks
  for food recognition,'' in \emph{WACV}, 2018, pp. 567--576.

\bibitem{Selvaraju2017Grad}
R.~R. Selvaraju, M.~Cogswell, A.~Das, R.~Vedantam, D.~Parikh, and D.~Batra,
  ``{Grad-CAM}: Visual explanations from deep networks via gradient-based
  localization.'' in \emph{ICCV}, 2017, pp. 618--626.

\bibitem{hassannejad2016food}
H.~Hassannejad, G.~Matrella, P.~Ciampolini, I.~De~Munari, M.~Mordonini, and
  S.~Cagnoni, ``Food image recognition using very deep convolutional
  networks,'' in \emph{MADiMaW}, 2016, pp. 41--49.

\bibitem{Hadsell-DR-CVPR06}
R.~Hadsell, S.~Chopra, and Y.~LeCun, ``Dimensionality reduction by learning an
  invariant mapping,'' in \emph{CVPR}, vol.~2, 2006, pp. 1735--1742.

\bibitem{Schroff-FaceNet-CVPR05}
F.~Schroff, D.~Kalenichenko, and J.~Philbin, ``Facenet: A unified embedding for
  face recognition and clustering,'' in \emph{CVPR}, 2015, pp. 815--823.

\bibitem{Radenovic-FTCNN-TPAMI2019}
F.~{Radenović}, G.~{Tolias}, and O.~{Chum}, ``Fine-tuning {CNN} image
  retrieval with no human annotation,'' \emph{IEEE TPAMI}, vol.~41, no.~7, pp.
  1655--1668, 2019.

\bibitem{weinberger2009distance}
K.~Q. Weinberger and L.~K. Saul, ``Distance metric learning for large margin
  nearest neighbor classification.'' \emph{JMLR}, vol.~10, no.~2, 2009.

\bibitem{carvalho2018cross}
M.~Carvalho, R.~Cad{\`e}ne, D.~Picard, L.~Soulier, N.~Thome, and M.~Cord,
  ``Cross-modal retrieval in the cooking context: Learning semantic text-image
  embeddings,'' in \emph{ACM SIGIR}, 2018, pp. 35--44.

\bibitem{liu2016ssd}
W.~Liu, D.~Anguelov, D.~Erhan, C.~Szegedy, S.~Reed, C.-Y. Fu, and A.~C. Berg,
  ``{SSD}: Single shot multibox detector,'' in \emph{ECCV}, 2016, pp. 21--37.

\bibitem{lin2017focal}
T.-Y. Lin, P.~Goyal, R.~Girshick, K.~He, and P.~Doll{\'a}r, ``Focal loss for
  dense object detection,'' in \emph{CVPR}, 2017, pp. 2980--2988.

\bibitem{ren2015faster}
S.~Ren, K.~He, R.~Girshick, and J.~Sun, ``{Faster R-CNN}: Towards real-time
  object detection with region proposal networks,'' in \emph{NIPS}, 2015, pp.
  91--99.

\bibitem{liu2018path}
S.~{Liu}, L.~{Qi}, H.~{Qin}, J.~{Shi}, and J.~{Jia}, ``Path aggregation network
  for instance segmentation,'' in \emph{CVPR}, 2018, pp. 8759--8768.

\bibitem{cai2018cascade}
Z.~Cai and N.~Vasconcelos, ``{Cascade R-CNN}: Delving into high quality object
  detection,'' in \emph{CVPR}, 2018, pp. 6154--6162.

\bibitem{zhang2020dynamic}
H.~Zhang, H.~Chang, B.~Ma, N.~Wang, and X.~Chen, ``{Dynamic R-CNN}: Towards
  high quality object detection via dynamic training,'' in \emph{ECCV}, 2020.

\bibitem{Ciocca2016Food}
G.~Ciocca, P.~Napoletano, and R.~Schettini, ``Food recognition: a new dataset,
  experiments, and results,'' \emph{IEEE JBHI}, vol.~21, no.~3, pp. 588--598,
  2016.

\bibitem{ziller2019oktoberfest}
A.~Ziller, J.~Hansjakob, V.~Rusinov, D.~Z{\"u}gner, P.~Vogel, and
  S.~G{\"u}nnemann, ``Oktoberfest food dataset,'' \emph{arXiv preprint
  arXiv:1912.05007}, 2019.

\bibitem{long2015fully}
J.~Long, E.~Shelhamer, and T.~Darrell, ``Fully convolutional networks for
  semantic segmentation,'' in \emph{CVPR}, 2015, pp. 3431--3440.

\bibitem{badrinarayanan2017segnet}
V.~Badrinarayanan, A.~Kendall, and R.~Cipolla, ``Segnet: A deep convolutional
  encoder-decoder architecture for image segmentation,'' \emph{IEEE TPAMI},
  vol.~39, no.~12, pp. 2481--2495, 2017.

\bibitem{zhao2017pyramid}
H.~Zhao, J.~Shi, X.~Qi, X.~Wang, and J.~Jia, ``Pyramid scene parsing network,''
  in \emph{CVPR}, 2017, pp. 2881--2890.

\bibitem{wang2018understanding}
P.~Wang, P.~Chen, Y.~Yuan, D.~Liu, Z.~Huang, X.~Hou, and G.~Cottrell,
  ``Understanding convolution for semantic segmentation,'' in \emph{WACV},
  2018, pp. 1451--1460.

\bibitem{peng2017large}
C.~Peng, X.~Zhang, G.~Yu, G.~Luo, and J.~Sun, ``Large kernel matters--improve
  semantic segmentation by global convolutional network,'' in \emph{CVPR},
  2017, pp. 4353--4361.

\bibitem{chen2018encoder}
L.-C. Chen, Y.~Zhu, G.~Papandreou, F.~Schroff, and H.~Adam, ``Encoder-decoder
  with atrous separable convolution for semantic image segmentation,'' in
  \emph{ECCV}, 2018, pp. 801--818.

\bibitem{dosovitskiy2020image}
A.~Dosovitskiy, L.~Beyer, A.~Kolesnikov, D.~Weissenborn, X.~Zhai,
  T.~Unterthiner, M.~Dehghani, M.~Minderer, G.~Heigold, S.~Gelly \emph{et~al.},
  ``An image is worth 16x16 words: Transformers for image recognition at
  scale,'' \emph{arXiv preprint arXiv:2010.11929}, 2020.

\bibitem{Furtado-HVS-CVIU2020}
P.~Furtado, M.~Caldeira, and P.~Martins, ``Human visual system vs convolution
  neural networks in food recognition task: An empirical comparison,''
  \emph{CVIU}, vol. 191, p. 102878, 2020.

\bibitem{Aoxue-LSFSL-CVPR2019}
A.~Li, T.~Luo, Z.~Lu, T.~Xiang, and L.~Wang, ``Large-scale few-shot learning:
  Knowledge transfer with class hierarchy,'' in \emph{CVPR}, 2019, pp.
  7212--7220.

\bibitem{goodfellow2014generative}
I.~Goodfellow, J.~Pouget-Abadie, M.~Mirza, B.~Xu, D.~Warde-Farley, S.~Ozair,
  A.~Courville, and Y.~Bengio, ``Generative adversarial nets,'' \emph{NIPS},
  vol.~27, 2014.

\bibitem{liu2020towards}
B.~Liu, Y.~Zhu, K.~Song, and A.~Elgammal, ``Towards faster and stabilized gan
  training for high-fidelity few-shot image synthesis,'' in \emph{ICLR}, 2020.

\bibitem{Kekai-Gourmet-SIGGRAPH2018}
K.~Sheng, W.~Dong, H.~Huang, C.~Ma, and B.-G. Hu, ``Gourmet photography dataset
  for aesthetic assessment of food images,'' in \emph{SIGGRAPH Asia}, 2018, pp.
  1--4.

\end{thebibliography}

\vspace{-35 pt}
\end{document}